\documentclass[10pt,twocolumn,letterpaper]{article}

\usepackage{multirow}

\usepackage{array}

\usepackage[pagenumbers]{iccv}
\definecolor{iccvblue}{rgb}{0.21,0.49,0.74}
\usepackage[pagebackref,breaklinks,colorlinks,allcolors=iccvblue]{hyperref}
\usepackage{bbding}

\title{A Self-Correcting Vision-Language-Action Model for Fast and Slow \\System Manipulation}

\author{
    \normalsize Chenxuan Li\textsuperscript{\rm 1$^{*}$}, Jiaming Liu\textsuperscript{\rm 1$^{*}$}, Guanqun Wang\textsuperscript{\rm 1$^{*}$}, Xiaoqi Li\textsuperscript{\rm 1},  Sixiang Chen\textsuperscript{\rm 1}, Liang Heng\textsuperscript{\rm 1}, Chuyan Xiong\textsuperscript{\rm 1},\\ 
    \normalsize Jiaxin Ge\textsuperscript{\rm 1}, Renrui Zhang\textsuperscript{\rm 1}, Kaichen Zhou\textsuperscript{\rm 1}, Shanghang Zhang\textsuperscript{\rm 1}~\textsuperscript{\Envelope}\\
    \normalsize \textsuperscript{\rm 1}State Key Laboratory of Multimedia Information Processing, School of Computer Science, Peking University
}

\begin{document}

\twocolumn[
{%
\renewcommand\twocolumn[1][]{#1}
\maketitle
\begin{center}
\centering
\begin{minipage}[t]{\linewidth}
\includegraphics[width=0.98\textwidth]{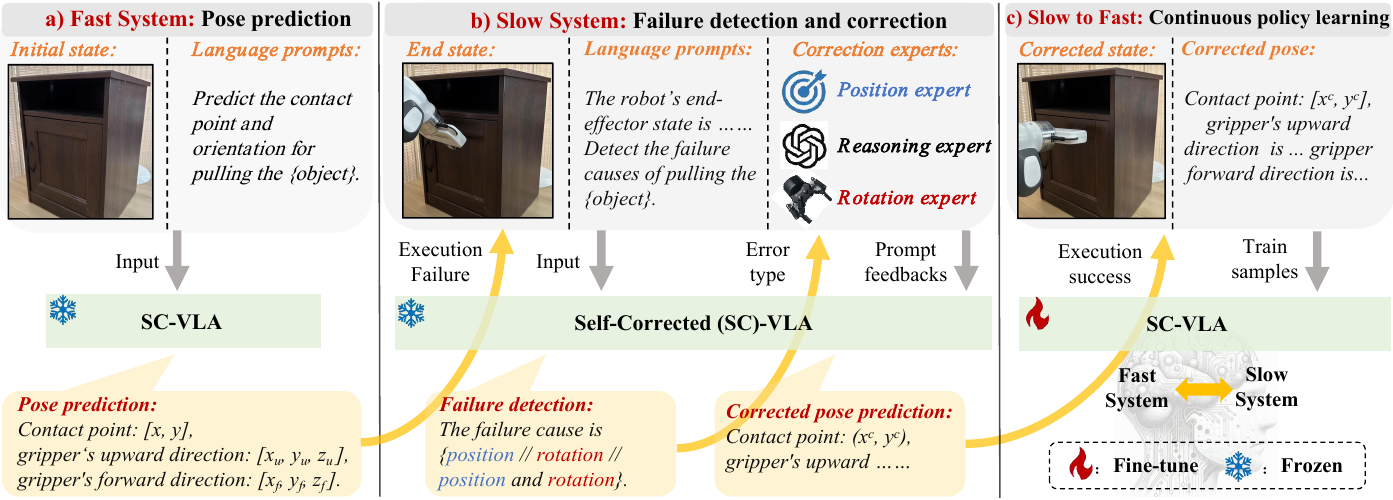}
\vspace{-0.2cm}
{\captionsetup{hypcap=false}  
\captionof{figure}{\footnotesize{
\textbf{Overview of self-corrected (SC)-VLA.}
a) \textbf{Fast system of SC-VLA} enhances the pretrained MLLM with pose prediction capabilities in an autoregressive manner.
b) \textbf{Slow system of SC-VLA} utilizes the end-state image and end-effector states to recognize failures, intelligently seeking expert prompt feedback to reflect on and correct the action pose through step-by-step reasoning.
c) \textbf{Slow to fast: Continuous policy learning.} 
When an action is successfully corrected, SC-VLA continuously learns policies from these samples, enhancing the manipulation accuracy of the fast system prediction.
}}
\label{fig:intro}}
\end{minipage}
\end{center}
}]

\maketitle
\renewcommand{\thefootnote}{}
\footnotetext{*: Equal contribution,   \Envelope: Corresponding author}

\begin{abstract}
Recently, some studies have integrated Multimodal Large Language Models into robotic manipulation, constructing vision-language-action models (VLAs) to interpret multimodal information and predict SE(3) poses. While VLAs have shown promising progress, they may suffer from failures when faced with novel and complex tasks. To emulate human-like reasoning for more robust manipulation, we propose the self-corrected (SC-)VLA framework, which integrates fast system for directly predicting actions and slow system for reflecting on failed actions within a single VLA policy. For the fast system, we incorporate parameter-efficient fine-tuning to equip the model with pose prediction capabilities while preserving the inherent reasoning abilities of MLLMs. For the slow system, we propose a Chain-of-Thought training strategy for failure correction, designed to mimic human reflection after a manipulation failure. Specifically, our model learns to identify the causes of action failures, adaptively seek expert feedback, reflect on the current failure scenario, and iteratively generate corrective actions, step by step. Furthermore, a continuous policy learning method is designed based on successfully corrected samples, enhancing the fast system's adaptability to the current configuration. We compare SC-VLA with the previous SOTA VLA in both simulation and real-world tasks, demonstrating an efficient correction process and improved manipulation accuracy on both seen and unseen tasks.

``System 1 is fast, automatic, and intuitive, operating with little to no effort. In contrast, System 2 is slow, deliberate, and conscious, requiring intentional effort.'' --- \cite{kahneman2011thinking}. 

\end{abstract}

\section{Introduction}
Recently, Multimodal Large Language Models (MLLMs)~\cite{li2022blip, alayrac2022flamingo, liu2023visual} have showcased remarkable abilities in common sense reasoning. 
Some studies~\cite{singh2023progprompt, ahn2022can, liang2023code, driess2023palm} integrate MLLMs into robot manipulation, enabling robots to explore multimodal information and formulate task planning. 
Concurrently, other researchers~\cite{zitkovich2023rt, li2023vision, li2023manipllm, kim2024openvla, liu2024robomamba} are focusing on developing vision-language-action (VLA) models for predicting robotic low-level action poses.
While the integration of MLLMs into robotics has made encouraging strides, the current pipelines may lead to failure predictions when faced with novel tasks or object instances. 
Most prior research overlooks the detection and correction of failure actions within the control process, limiting their practicality in real-world scenarios where uncertainties and unexpected obstacles are common.

Recognizing the crucial role of correction in manipulation, recent studies have proposed solutions.
REFLECT~\cite{liu2023reflect} stands out by utilizing LLMs to generate failure explanations and assist a language-based planner in correcting errors. 
Building on this innovation, subsequent research~\cite{guo2023doremi, skreta2023errors, shi2024yell, zha2023distilling, ming2023hicrisp, yu2023multireact} delve deeper into LLMs' robot correction capabilities, reflecting on failure scenes and generating correction prompts (e.g.,``Move a little bit to the right").
However, the existing correction methods still face two main challenges.
\textbf{1) Lack of ability to directly correct low-level SE(3) poses.} 
Existing methods struggle to correct end-effector poses, as language correction prompts often cause misinterpretations in the action model due to its lack of prior exposure during training.
\textbf{2) Lack of ability to learn from correction feedback.} Current correction frameworks fail to learn from successful correction cases and improve policy predictions. They still rely on external experts (e.g., GPT-4~\cite{achiam2023gpt}) when encountering similar failures.
Given these challenges, we raise a question: ``\textit{Can we develop a VLA model that not only possesses manipulation skills but also effectively corrects low-level failures?”}

Drawing inspiration from Daniel Kahneman's assertion that ``human thinking is divided into a fast system and a slow system, which separately represent intuitive processes and more logical reasoning,"~\cite{kahneman2011thinking} we introduce a self-corrected (SC-) VLA framework that mimics a human-like thinking paradigm to address the above question. 
As shown in Figure~\ref{fig:intro}, the SC-VLA possesses both the fast system capability to directly predict end-effector poses and the slow system ability to reflect on and correct failure actions within a single VLA model.
For fast system construction, we transform low-level actions into a language modeling problem, adopting parameter-efficient fine-tuning to the LLM while preserving its inherent reasoning abilities.
To equip the SC-VLA with slow system capabilities, we propose a Chain-of-Thought training strategy for failure correction, designed to emulate human reflection when encountering manipulation failures.
Specifically, the SC-VLA first recognizes SE(3) pose errors as position, rotation, or combined errors by utilizing the final state image and the robot state. Based on the identified error type, the SC-VLA adaptively requests correction feedback from experts, such as position~\cite{mo2021where2act}, rotation~\cite{ fang2023anygrasp}, and reasoning expert~\cite{achiam2023gpt}. Drawing on insights from previous errors and expert prompts, the SC-VLA reflects on the current failure scenario and regenerates the actions.

During inference, the SC-VLA initially interacts with the physical world using the fast system, activating the slow system for reflection and correction only when a failure action occurs. 
The transition between the two systems in SC-VLA can be flexibly switched using different input question prompts.
Moreover, as shown in Figure~\ref{fig:intro} c), we design a continuous policy learning method that transfers successfully corrected samples from the slow system to the fast system, using exponential moving average techniques~\cite{tarvainen2017mean} along with injected adapters. This approach enhances the model’s manipulation stability in the current scene configuration while reducing the frequency of expert intervention.
To train our SC-VLA, we generate 12k manipulation success samples, 15k failure samples, and 60k correction prompts in the SAPIEN simulation~\cite{Xiang_2020_SAPIEN}. We evaluate our method through extensive experiments in both simulation and real-world datasets. In SAPIEN, SC-VLA improves the manipulation success rate from 66\% to 87\% in seen tasks and from 30\% to 68\% in unseen tasks. 
In real-world experiments, SC-VLA achieves sim-to-real transfer through simple fine-tuning on novel manipulation demonstrations with fast system training prompts, delivering superior performance compared to previous methods.
In summary, our contributions are as follows:

\textbf{1)} To mimic a human-like thinking paradigm in manipulation, we propose a Self-Corrected (SC)-VLA, equipping our model with not only the fast system ability to predict end-effector poses but also the slow system ability to reflect on and correct failure actions.

\textbf{2)} For the slow system, we propose a Chain-of-Thought training strategy for detecting, reflecting on, and correcting low-level failure actions. SC-VLA can adaptively request expert prompt feedback to regenerate the contact pose.

\textbf{3)} For successfully corrected samples, we introduce a continuous policy learning method to progressively enhance the model's stability in the current scene configurations while reducing the frequency of expert intervention.

\vspace{-8pt}
\label{sec:intro}

\section{Related Work}

\textbf{Robotic Manipulation.}
Robotic manipulation is a crucial research area with broad applications. State-based reinforcement learning is widely used~\cite{joshi2020robotic, andrychowicz2020learning, yarats2021mastering, geng2022end}. While some works use pure state as policy input~\cite{andrychowicz2020learning}, complex tasks often require vision-based observation~\cite{Mo_2019_CVPR, mo2021where2act, liu2024rgbgrasp, xu2022universal, eisner2022flowbot3d, wu2021vat, huang2023voxposer, zitkovich2023rt, xu2023unidexgrasp, wan2023unidexgrasp++, gong2023arnold, yang2023equivact, wang2023sparsedff, geng2023sage} for environment perception~\cite{deng2023banana, lei2023nap}.
Inspired by MLLMs' success in general tasks, several works apply MLLMs' reasoning capabilities to manipulation. Palm-E~\cite{driess2023palm} trains multimodal encodings with LLMs for manipulation planning. VoxPoser~\cite{huang2023voxposer} uses MLLMs for zero-shot trajectory generation. RT-2~\cite{zitkovich2023rt} enables fast adaptation to novel tasks. Robotflamingo~\cite{li2023vision} fine-tunes MLLMs on imitation learning tasks for long-horizon manipulation. Recent approaches~\cite{li2023manipllm, liu2024robomamba} enhance MLLMs with end-effector pose prediction.

\textbf{Robotic Failure Correction.}
Several studies focus on correcting robotic failures. REFLECT~\cite{liu2023reflect} uses LLMs for reasoning based on past experiences, improving planning with failure explanations. MULTIREACT~\cite{yu2023multireact} applies a vision-language model~\cite{radford2021learning} as a reward model for recognizing and recovering from intermediate failures. DoReMi~\cite{guo2023doremi} detects misalignments between plans and execution in real-time using LLMs. CLAIRify~\cite{skreta2023errors} ensures plan validity through iterative prompting. Additional related work is provided in Appendix.

\label{sec:related}

\section{Approach}
\label{sec:approach}
In Section \ref{sec:PF}, we introduce the design motivation and problem formulation of SC-VLA. Subsequently, in Section \ref{sec:MCC}, we present the proposed self-corrected vision-language-action (SC-VLA) model, detailing how the model is equipped with both fast system and slow system capabilities. Finally, we explain the continuous policy learning mechanism in Section \ref{sec:CPL}.
\begin{figure*}[t]
\begin{center}
   \includegraphics[width=0.95\textwidth]{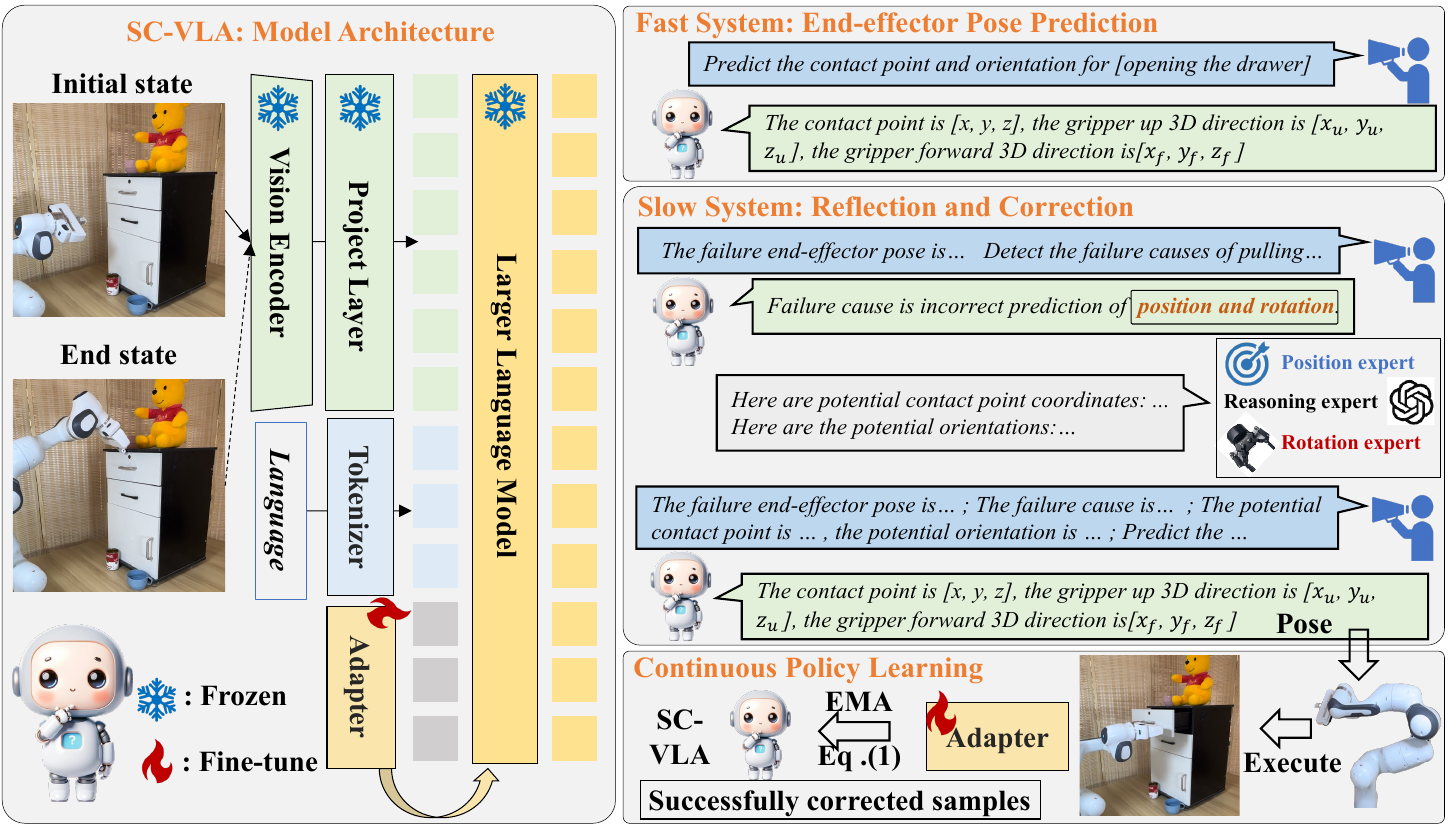}
\vspace{-0.2cm}
\caption{
\textbf{Overall framework of SC-VLA.} 
The left part shows the SC-VLA model architecture, where a vision encoder and projection layer transform robot images into LLM language embeddings. These embeddings are combined with text tokens and input into the LLM to generate end-effector poses. The right part illustrates SC-VLA's training strategy, using both the fast and slow systems. The fast system predicts poses, while the slow system corrects failures through CoT reasoning strategy. SC-VLA uses the same prompt format for both training and inference. After correcting failures, continuous policy learning is designed to fine-tune the adapters to improve the fast system’s manipulation stability.
}
   \label{fig:frame}
\end{center}
\vspace{-0.6cm}
\end{figure*}

\subsection{Overview}
\label{sec:PF}

Inspired by Kahneman’s dual-process theory \cite{kahneman2011thinking}, which posits that human cognition operates through two distinct systems—the fast system and the slow system—we propose the self-corrected vision-language-action (SC-VLA) framework. In this theory, the fast system is automatic and intuitive, enabling quick decisions based on prior experience. In contrast, the slow system is deliberate and reflective, involving complex reasoning for challenging situations.
Building upon this, SC-VLA integrates both systems to optimize the robotic manipulation process. 
The fast system enables direct end-effector pose prediction, allowing the robot to make rapid decisions.

However, relying solely on quick, intuitive responses can lead to failure actions, especially in complex or unexpected scenarios. To address this, the slow system is employed to reflect on and correct low-level SE(3) failures through our designed Chain-of-Thought training strategy.
The integration of fast and slow systems in SC-VLA mimics the human manipulation process, where rapid pose prediction and reflection work in tandem, improving the robustness of robotic actions.
Next, we describe the problem formulation of our model.

\textbf{End-effector Pose Prediction.}
The core task of SC-VLA is to predict the end-effector pose. Our SC-VLA policy, denoted as $\pi$, generates an action $a_i$ based on the initial input image ($I_i \in \mathbb{R}^{W \times H \times 3}$) and the corresponding language prompt ($L_i$). This action corresponds to the 6-DoF control of a Franka Panda robot arm, where $a_i = (a^\mathrm{pos}_i, a^\mathrm{rot}_i)$. Here, $a^\mathrm{pos}_i \in \mathbb{R}^3$ represents the 3D position of the end-effector, and $a^\mathrm{rot}_i \in \mathbb{R}^{3 \times 3}$ is the rotation matrix that describes the orientation. Following~\cite{li2023manipllm, kim2024openvla}, the end-effector pose is embedded directly into the language prompt during fine-tuning, enabling the model to predict actions along with textual information in an autoregressive manner.

\textbf{Failure Reflection and Correction.} 
In practice, the robot action $a_i$ often encounters failures, which are represented as error actions $a^{err}_i$. For failure correction, our SC-VLA policy utilizes the end-state RGB image ($I_e \in \mathbb{R}^{W \times H \times 3}$) and the language-descriptive robot state ($L^{a_i}_i$) to identify failure cases. This is formulated as $\pi(I_e, L^{a_i}_i) \rightarrow c_i$, where $c_i$ denotes the failure type (e.g., position or rotation error). 
Based on $c_i$, the slow system dynamically requests corrective feedback $f_i$ from experts. The feedback is then fed into the SC-VLA policy to re-predict the action $a_c$ using the adjusted input $\pi(I_i, f_i) \rightarrow a_c$.

\subsection{Self-Corrected VLA}
\label{sec:MCC}

To equip our model with foundational reasoning abilities for robotic manipulation, we load pretrained parameters from LLaMA-Adapter V2~\cite{gao2023llama}. 
This choice provides flexibility, as our approach can easily accommodate other MLLMs as the base model. As illustrated in Figure \ref{fig:frame}, the model architecture is composed of several key components. 
Due to space constraints, detailed architectural descriptions are provided in the Appendix. Next, we explain how we equip the base model with both fast and slow system capabilities, allowing it to make quick decisions while robustly refining its actions with corrective feedback.

\subsubsection{Fast system: pose prediction}
The fast system of our SC-VLA model is designed to directly generate end-effector poses for robotic manipulation tasks. During the pre-collection of training data, we capture RGB images and their corresponding successful end-effector poses for imitation learning.
During fine-tuning, as shown in the fast system part of Figure \ref{fig:frame}, we structure the input text prompt for pose prediction as: ``\textit{Predict the contact point and orientation for [task name].}'' The output format is specified as: ``\textit{The contact point is [$x$, $y$, $z$], the gripper up 3D direction is [$x_u$, $y_u$, $z_u$], and the gripper forward 3D direction is [$x_f$, $y_f$, $z_f$].}'' 
To represent the 6-DOF pose in text, we convert it into a classification problem by discretizing the continuous values into bins ranging from (-50, 50] using 100 normalized integer vectors~\cite{li2023manipllm, zitkovich2023rt}. 
For the trajectory prediction, we continuously feed the current state image into the model to generate the next 6-Dof pose.

\subsubsection{Slow system: reflection and correction}
\label{subsec:FDC}

The slow system in our SC-VLA model is designed to reflect on and correct failure actions during manipulation. Unlike previous works~\cite{liu2023reflect, ming2023hicrisp} that focus on correcting high-level planning, we introduce the novel approach of directly correcting the end-effector's 6-DoF control actions through a Chain-of-Thought (CoT) strategy. This correction operation is applied to key frames in the trajectory, which represent crucial or bottleneck steps of the gripper during task execution~\cite{goyal2023rvt, shridhar2023perceiver}, such as pre-pick, grasp, or place poses.
Before performing the correction, the system first assesses whether the action has failed and identifies the cause of the failure. Leveraging our model's visual understanding capabilities, we use specific prompts to determine task completion. For example, we input the end-state image and ask, ``\textit{Is the drawer open? Please answer yes or no.}'' If the task is incomplete, the system reflects on the failure case by breaking it down into potential causes, such as incorrect position or rotation. This step-by-step reasoning approach, inspired by CoT, allows the model to systematically identify and address the underlying issues.

\begin{figure*}[t]
\begin{center}
   \includegraphics[width=0.9\textwidth]{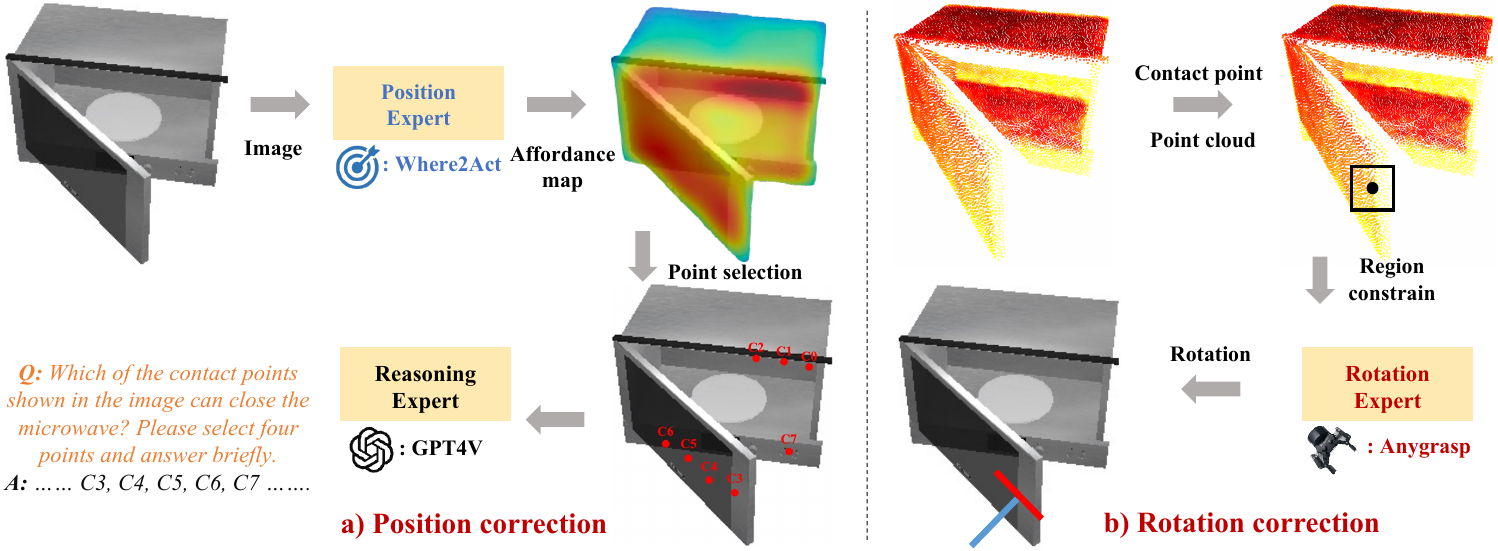}
\vspace{-0.3cm}
\caption{
\textbf{The process of expert feedback.} 
\textcolor{red}{For position correction}, we sequentially use the position expert and reasoning expert to generate an affordance map and select contact points for prediction. \textcolor{red}{For rotation correction}, we automatically identify the manipulation area and leverage the rotation expert to predict potential action rotations.
}
   \label{fig:cor}
\end{center}
\vspace{-0.4cm}
\end{figure*}

Since the pre-trained MLLM lacks failure recognition capabilities, we collected 15,000 failure samples and 60,000 correction prompts in the SAPIEN simulator~\cite{Xiang_2020_SAPIEN}. These failures are categorized into position, rotation, and combined errors. For position errors, we check if the contact point falls within the object's affordance region. For rotation errors, we assess whether the angle between the predicted Z-axis and the normal of the object's movable plane exceeds 30 degrees. Both the affordance region and the movable parts are automatically generated in the simulator, enabling rapid failure analysis. Once the failure cause is identified, we fine-tune our model's adapter using failure detection prompts, as shown in Figure \ref{fig:frame}. Based on the error type, SC-VLA dynamically requests corrective feedback from specific experts, as illustrated in Figure \ref{fig:cor}. For position correction, we use Where2Act~\cite{mo2021where2act} to generate an affordance map from the input image, indicating potential manipulable regions. However, due to the limitations of existing affordance-based methods~\cite{Mo_2019_CVPR, eisner2022flowbot3d}, we further refine the predicted contact points using a reasoning expert (e.g., GPT-4V~\cite{achiam2023gpt}). We select points with high affordance scores, project them back onto the image, and ask the reasoning expert, ``\textit{Which of the contact points shown in the image can close the object? Please select n points.}'' This iterative refinement ensures more reliable contact points.
For rotation correction, as shown in Figure \ref{fig:cor} b), we use Anygrasp~\cite{fang2023anygrasp} to predict potential rotations within a manipulation box. The manipulation box is defined by expanding 5 pixels outward from the predicted contact point. The rotation expert uses the highest-confidence contact point provided by the position expert to generate the corresponding rotation. If the failure involves both position and rotation errors, the system applies position correction first, followed by rotation correction.
To encourage SC-VLA to reflect on failure actions through CoT reasoning, we combine the erroneous action's 6DOF pose, the cause of the error, and the expert's correction feedback, using them as input prompts for our model to re-predict the pose, as shown in Figure \ref{fig:frame}.
This correction CoT process mimics human manipulation reasoning, where we often analyze errors step by step—first recognizing what went wrong, then understanding the underlying cause, and finally adjusting our actions based on expert guidance or prior knowledge.
By adopting this slow-system reasoning, we enable SC-VLA to break down the failure action into manageable components, such as identifying where the error occurred (pose), why it happened (cause), and how it can be rectified (correction feedback). This design significantly improves the accuracy of actions for failure samples, as demonstrated in Section~\ref{sec:MMS} and Section~\ref{sec:RWE}.

\subsection{Continuous policy learning}
\label{sec:CPL}

To equip our model with both the fast system's ability for direct manipulation and the slow system's ability for failure correction, we integrate pose prediction, failure detection, and failure correction data for co-fine-tuning using cross-entropy loss.
During inference, the relative position of the robot and the object changes with each execution, making it impractical to reuse previous expert feedback. To address this, once we obtain successfully corrected samples, we implement a continuous policy learning (CPL) method, as illustrated in Figure \ref{fig:frame}. This method enhances the fast system's pose prediction accuracy without relying on the slow system's CoT reasoning and expert interventions.

In continuous policy learning, we update the model after every fifty corrected samples, using the same prompt format as in the fast system training. However, fine-tuning on new successful samples risks catastrophic forgetting, potentially reducing accuracy on previously trained samples.
To mitigate this risk, we employ an exponential moving average (EMA)~\cite{tarvainen2017mean} along with a parameter-efficient fine-tuning strategy, enabling continuous learning from new data while preserving previously learned knowledge. The EMA update rule is $\mathcal{\pi}^{t} = \alpha \mathcal{\pi}^{t-1} + (1-\alpha) \mathcal{\pi}^{t}$.
\label{eq:ema}Where $t$ represents the time step, and $\pi$ denotes our model. 
We set the updating weight $\alpha = 0.999$ based on \cite{AnttiTarvainenetal2017}. The effectiveness of this EMA scheme for action continual learning is evaluated in the Appendix.
In summary, SC-VLA can iteratively perform continuous policy learning using corrected samples, progressively transferring knowledge from the slow system to the fast system. This mechanism enables the model to adapt to environmental variations over time, without relying solely on rule-based corrections.

\begin{table*}[t]
    \centering
    \small
    \setlength{\tabcolsep}{2pt}
    \renewcommand{\arraystretch}{1.2}
    \caption{
   Comparison of SC-VLA with baseline methods. The table shows the performance of different methods across various seen and unseen tasks.
    ``Experts" refers to our position and rotation experts generating action poses in a zero-shot manner.
    ``Fast" and ``Slow" represent our method's results for the fast system's direct prediction and the slow system's CoT corrected prediction based on expert prompts, respectively. ``CPL" refers to using our continuous policy learning method. Each task icon is described in the Appendix.
    }
    \label{tab:main}

    \resizebox{\textwidth}{!}{
    \begin{tabular}{c|cccccccccccccccc}
        \hline
        \multirow{2}{*}{\textbf{\raisebox{-1.0\height}{Method}}} & \multicolumn{16}{c}{\textbf{Seen tasks}} \\
        \cline{2-17}
        & \includegraphics[width=0.050\linewidth]{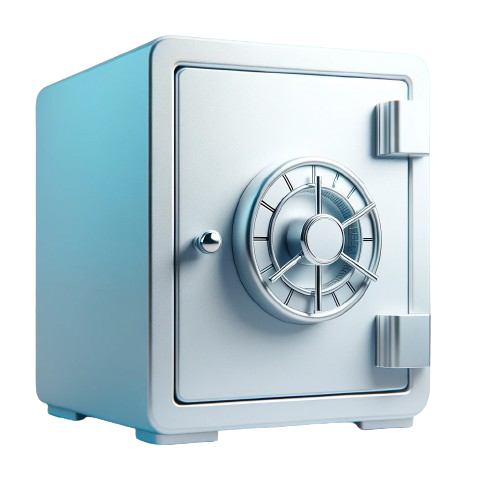}
        &\includegraphics[width=0.050\linewidth]{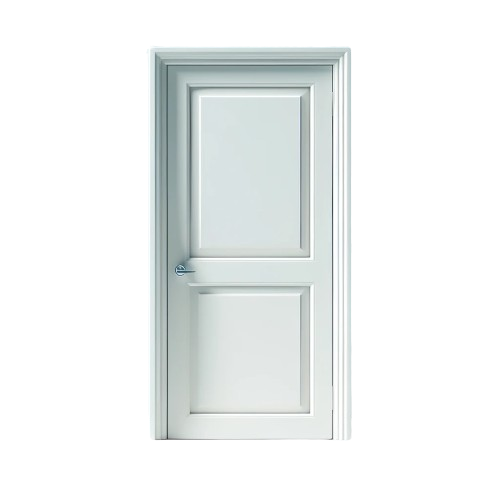}
        &\includegraphics[width=0.050\linewidth]{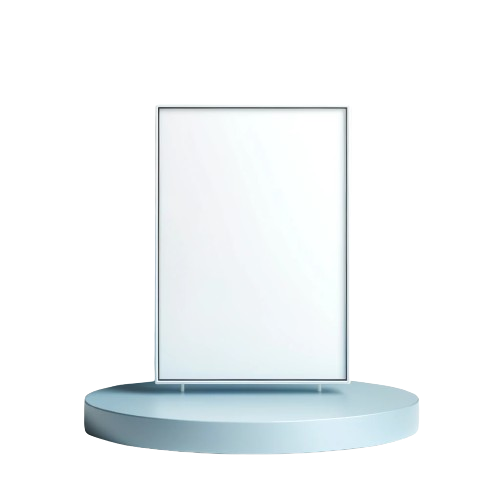}
        &\includegraphics[width=0.050\linewidth]{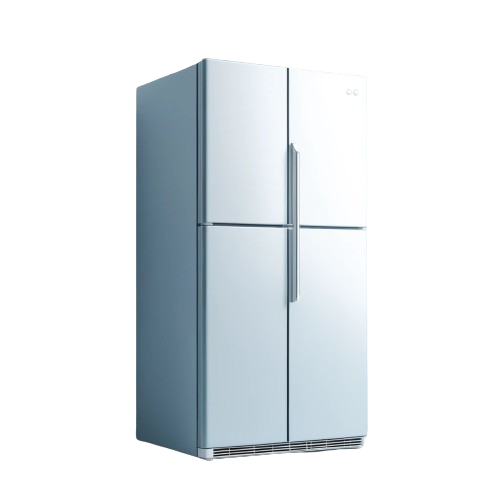}
        &\includegraphics[width=0.050\linewidth]{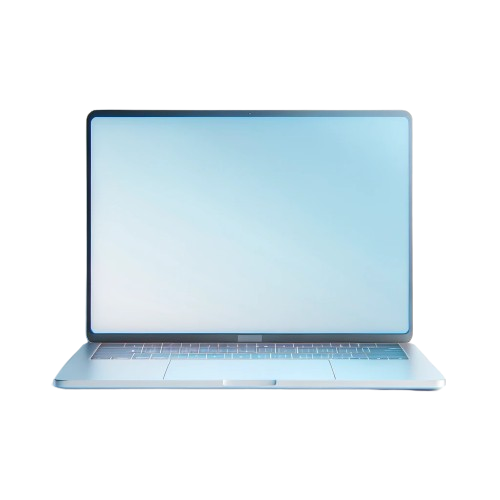}
        &\includegraphics[width=0.050\linewidth]{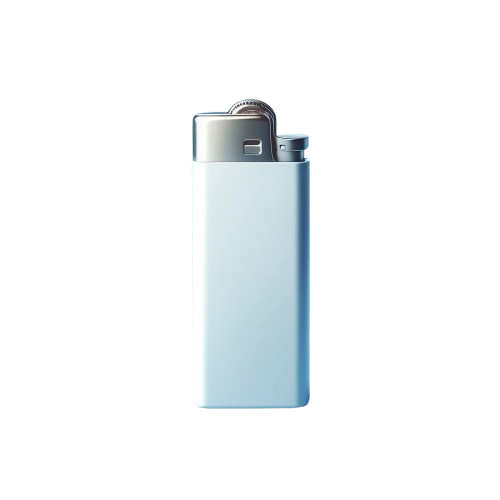}
        &\includegraphics[width=0.050\linewidth]{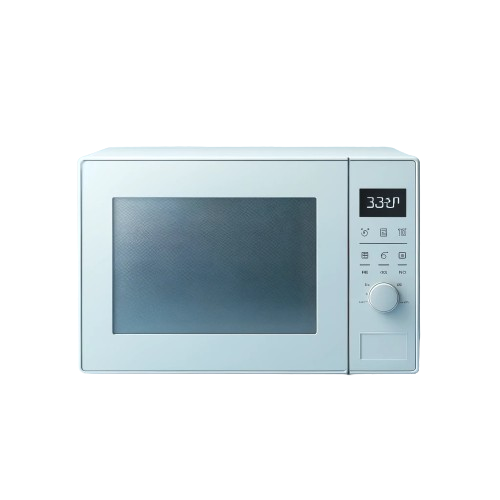}
        &\includegraphics[width=0.050\linewidth]{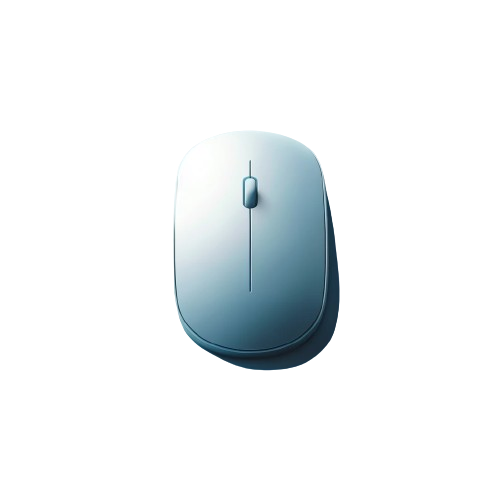}
        &\includegraphics[width=0.050\linewidth]{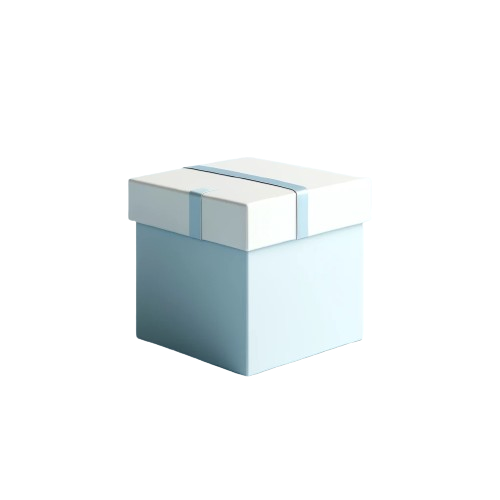}
        &\includegraphics[width=0.050\linewidth]{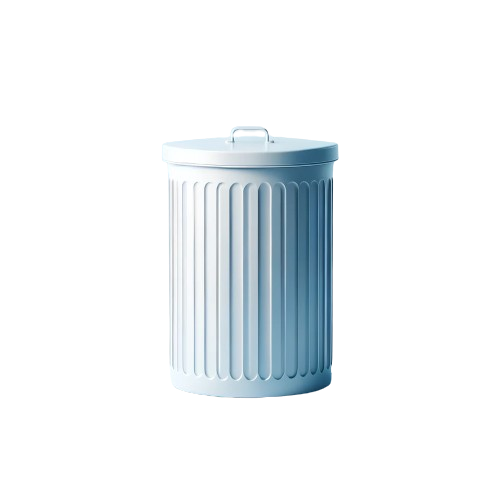}
        &\includegraphics[width=0.050\linewidth]{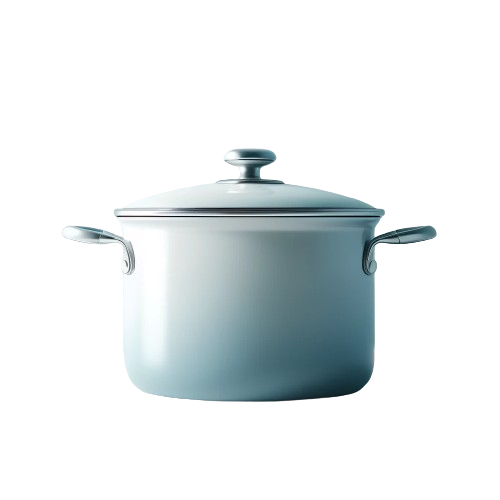}
        &\includegraphics[width=0.050\linewidth]{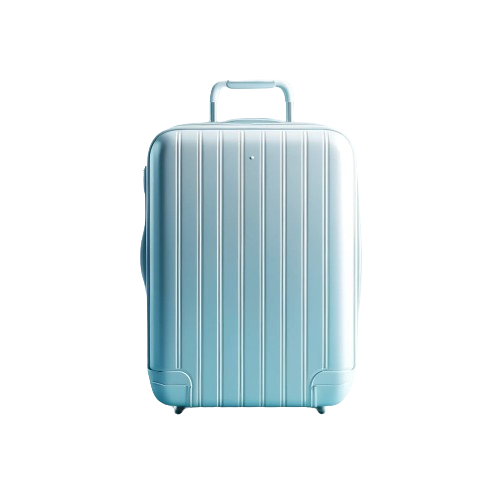}
        &\includegraphics[width=0.050\linewidth]{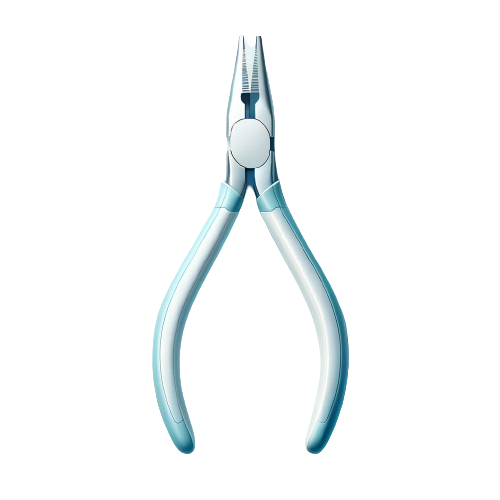}
        &\includegraphics[width=0.050\linewidth]{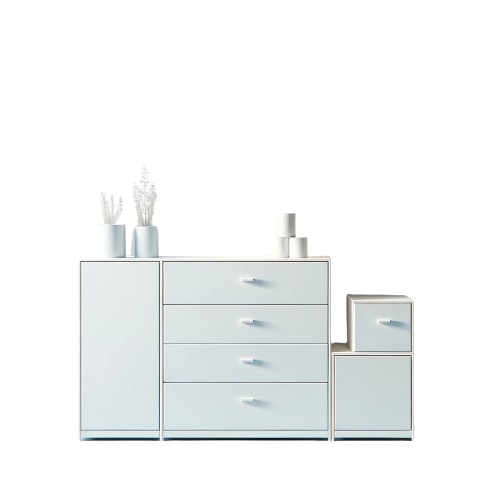}
        &\includegraphics[width=0.050\linewidth]{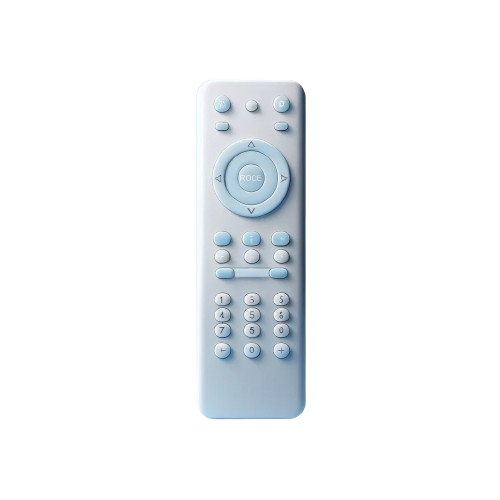}
        &\includegraphics[width=0.050\linewidth]{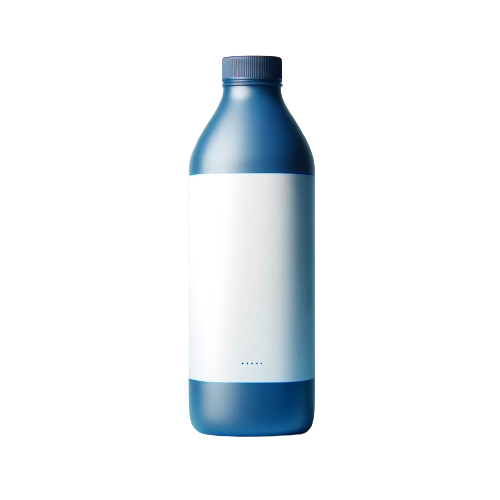} \\
        \hline\hline 
        UMPNet~\cite{xu2022universal} &0.23	&0.36	&0.41	&0.22	&0.24&	0.30	&0.43	&0.34&	\textbf{0.51}	&0.21	&0.66	&0.27	&0.23	&0.23	&0.29	&0.60\\
        FlowBot3D~\cite{eisner2022flowbot3d} &0.45&	0.48&	0.45&	0.32&	0.32&	0.37&	0.43&	0.23&	0.26&	0.14&	0.39&	0.31&	0.38&	0.32&	0.23&	0.43 \\
        ManipLLM~\cite{li2023manipllm} &0.72	&0.56	&0.32	&0.79	&0.48	&0.53	&0.66	&0.69	&0.39	&0.52	&0.53	&0.4	&\textbf{0.64}	&0.73	&\textbf{0.62}	&0.52 \\
        Experts& 0.34	&0.36	&0.33&	0.44&	0.45&	\textbf{0.56}&	0.32&	0.19&	0.48&	0.28&	0.53&0.29&	0.27&	0.32&	0.27&	0.45	 \\
        \hline
        Ours(Fast) & 0.78 & 0.63 & 0.58 & 0.70 & 0.52 & 0.13 & 0.81 & 0.88 & 0.56 & 0.71 & 0.84 & 0.80 & 0.46 & 0.76 & 0.30 & 0.83 \\ 
        \hline 
        Ours(Fast+Slow) & 0.97 & 0.90 & 0.66 & 0.93 & 0.95 & 0.66 & 0.97 & 0.96 & 0.87 & 0.92 & 0.90 & 0.87 & 0.78 & 0.94 & 0.30 & 0.90 \\
        \hline
        Ours(CPL+Fast) & \textbf{0.90} & \textbf{0.75} & \textbf{0.58} & \textbf{0.87} & \textbf{0.95} & 0.46 & \textbf{0.89} & \textbf{0.92} & 0.50 & \textbf{0.78} & \textbf{0.90} & \textbf{0.85} & 0.63 & \textbf{0.90} & 0.38 & \textbf{0.90} \\
        \hline
    \end{tabular}
    }
    \vspace{0.2cm}
    \resizebox{\textwidth}{!}{
    \begin{tabular}{c|ccccc|ccccccccccc}
        \hline
        \multirow{2}{*}{\textbf{\raisebox{-1.0\height}{Method}}} & \multicolumn{5}{c|}{\textbf{Seen tasks}} & \multicolumn{11}{c}{\textbf{Unseen tasks}}\\
        \cline{2-17}
        & \includegraphics[width=0.050\linewidth]{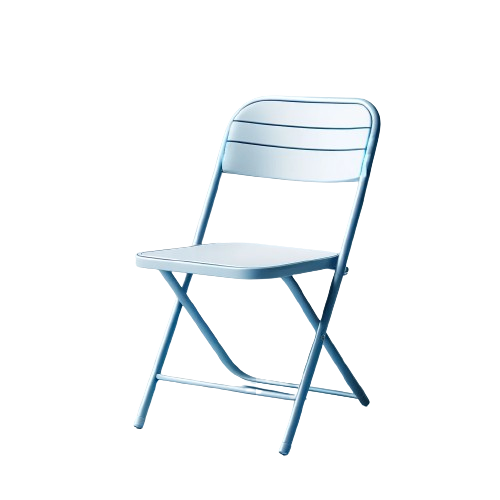}
        &\includegraphics[width=0.050\linewidth]{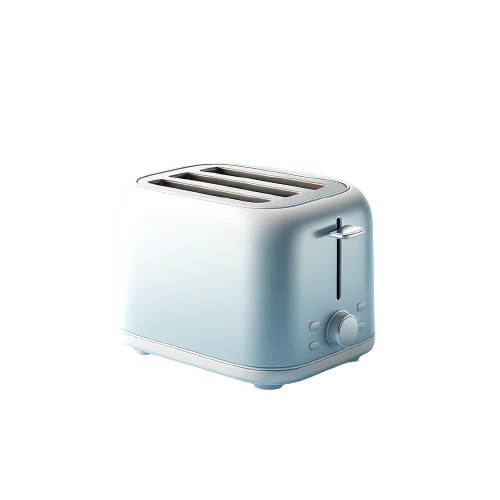}
        &\includegraphics[width=0.050\linewidth]{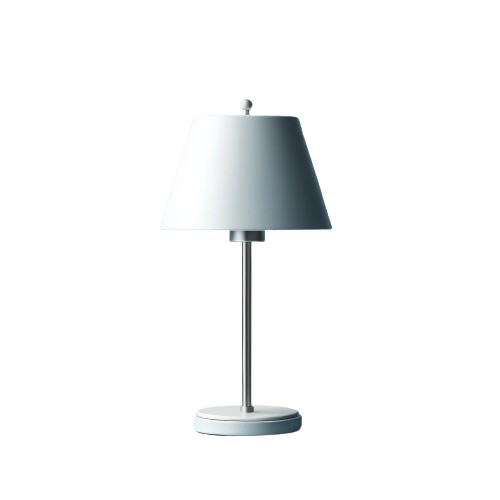}
        &\includegraphics[width=0.050\linewidth]{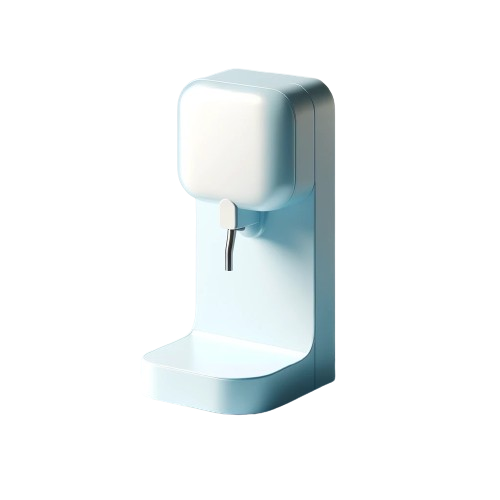}
        &\multicolumn{1}{c|}{{\textbf {\raisebox{0.5\height}{AVG}}} }
        &\includegraphics[width=0.050\linewidth]{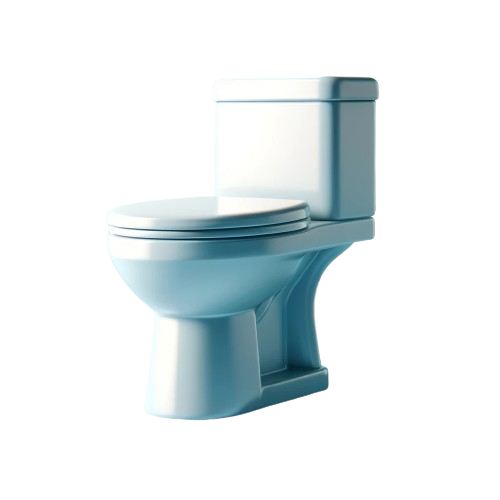}
        &\includegraphics[width=0.050\linewidth]{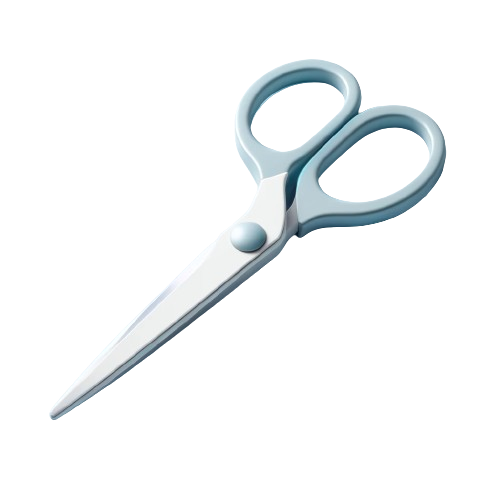}
        &\includegraphics[width=0.050\linewidth]{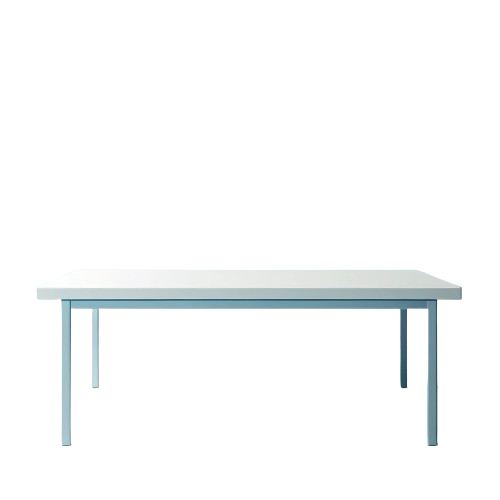}
        &\includegraphics[width=0.050\linewidth]{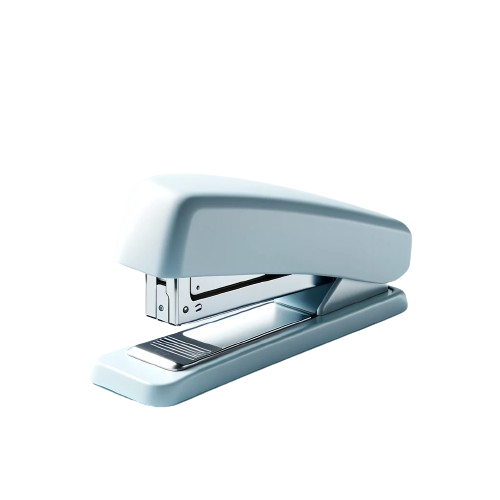}
        &\includegraphics[width=0.050\linewidth]{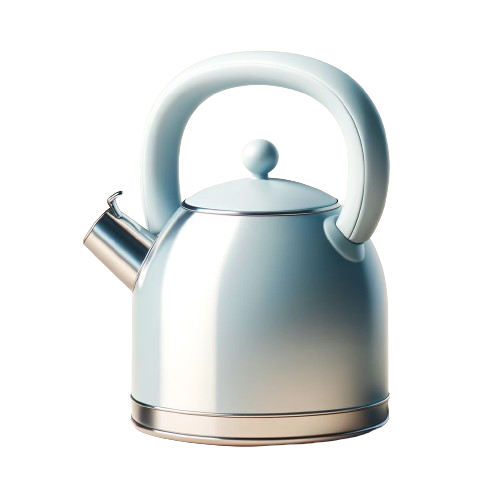}
        &\includegraphics[width=0.050\linewidth]{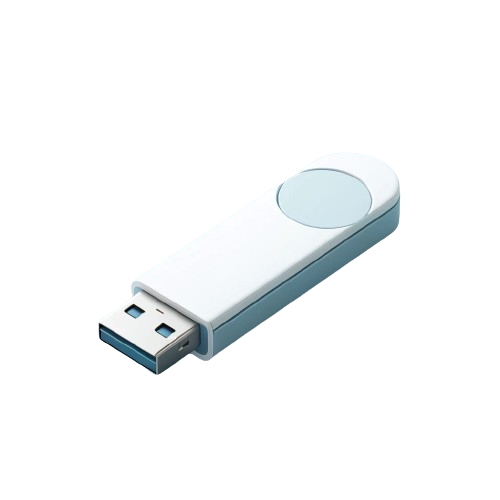}
        &\includegraphics[width=0.050\linewidth]{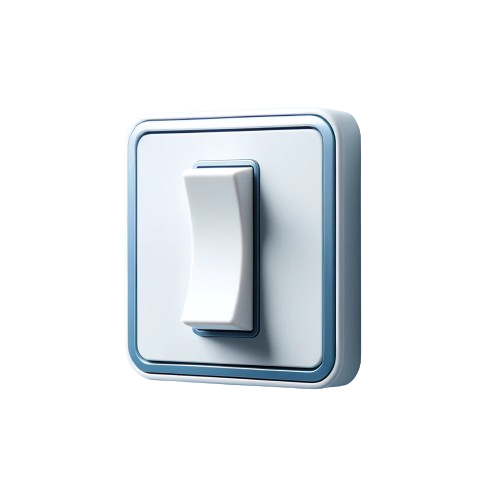}
        &\includegraphics[width=0.050\linewidth]{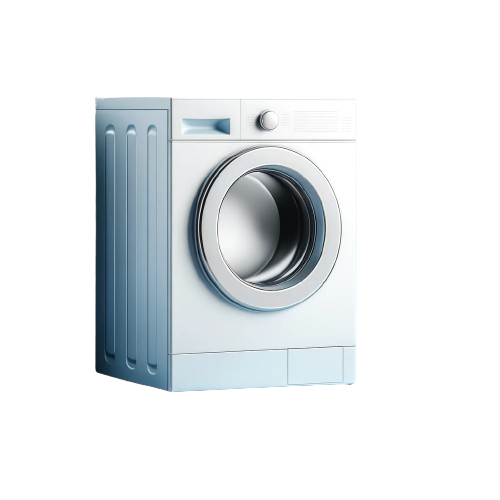}
        &\includegraphics[width=0.050\linewidth]{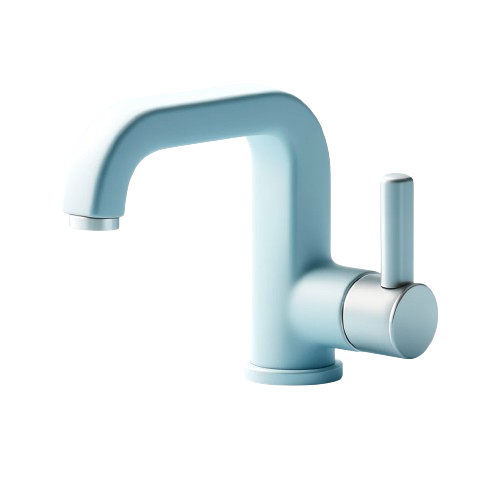}
        &\includegraphics[width=0.050\linewidth]{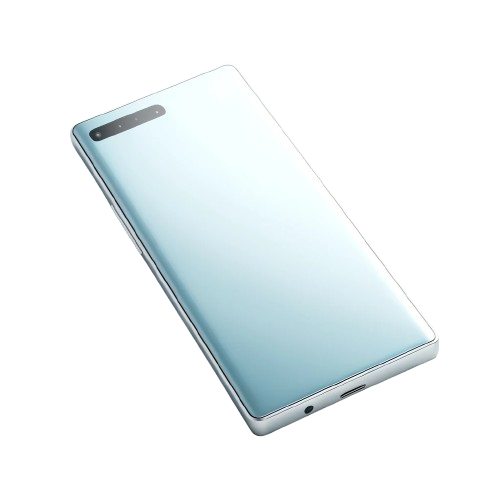}
        &{\textbf {\raisebox{0.5\height}{AVG}}} \\
        \hline\hline 
        UMPNet~\cite{xu2022universal}& 0.32&	0.30&	0.11&	0.58&  0.34& \textbf{0.36}&	0.36&	0.38&	0.47&	0.21&	0.12&	0.24&	0.23&	0.28&	0.12&	0.28  \\
        FlowBot3D~\cite{eisner2022flowbot3d} &0.19&	0.33&	0.23&	0.47&	0.33& 0.29&	0.47&	0.64&	0.31&	0.27&	0.30&	0.09&	0.41&	0.35&	0.37&	0.35\\
        ManipLLM~\cite{li2023manipllm} &0.39	&\textbf{0.75}	&0.44	&0.67	&0.57 &0.32	&0.22	&0.65	&\textbf{0.69}	&0.38	&0.85	&0.27	&0.53	&0.26	&0.38	&0.47\\
        Experts &0.21&	0.49&	0.29	&0.24	&0.36& 0.33&	0.36	&0.49&	0.36&	0.19&	0.42&	0.28&	0.41	&0.47&	0.56	&0.39 \\
        \hline
        Ours(Fast) & 0.20 & 0.68 & 0.48 & 0.60 & 0.66 &0.09 & 0.25 & 0.39 & 0.66 & 0.64 & 0.23 & 0.21 & 0.56 & 0.10 & 0.50 & 0.30 \\ 
        \hline 
        Ours(Fast+Slow) & 0.80 & 0.89 & 0.74 & 0.98 & 0.87 & 0.27 & 0.65 & 0.71 & 0.83 & 0.85 & 0.71 & 0.73 & 0.87 & 0.48 & 0.90 & 0.68 \\
        \hline
        Ours(CPL+Fast) & \textbf{0.60} & 0.71 & \textbf{0.74} & \textbf{0.90} & \textbf{0.79} & 0.27 & \textbf{0.61} & \textbf{0.71} & 0.50 & \textbf{0.92} & \textbf{0.69} & \textbf{0.69} & \textbf{0.80} & \textbf{0.70} & \textbf{0.81} & \textbf{0.69} \\
        \hline
    \end{tabular}
    }
    \label{tab:combined}
\end{table*}

\section{Experiment}
In this section, we conduct extensive experiments in both simulation and real-world settings. First, we introduce the simulation experimental setup in Sec.\ref{sec:ES}. In Sec.\ref{sec:MMS}, we compare our SC-VLA with previous baselines on the simulation dataset. The ablation study is presented in Sec.\ref{sec:AS}, demonstrating the effectiveness of each component. Finally, in Sec.\ref{sec:RWE}, we present the robust manipulation ability of our SC-VLA in real-world scenarios.

\subsection{Simulation experiment setting}
\label{sec:ES}
In our experiments, we evaluate both open-loop and close-loop control for robotic manipulation tasks. For the open-loop experiment, we use the SAPIEN~\cite{Xiang_2020_SAPIEN} engine to simulate interactions with objects from PartNet-Mobility~\cite{mo2019partnet}, collecting a dataset of 12k successful manipulation samples and 15k failure samples. 
To prevent the potential sim-to-real gap, in simulation training, we increase data diversity by varying object poses, camera angles, and lighting. 
We evaluate the model's generalization using a test set with both seen and unseen categories. 
The close-loop experiments are conducted using RLBench tasks~\cite{james2020rlbench} with a single front view input and predefined key frame selection~\cite{shridhar2023perceiver,goyal2023rvt}. 
We collect 100 episodes per task as the training dataset and conduct 20 test trials in the simulator for each task.
We use the Adam optimizer with a learning rate of 2e-5 to train the model on 80GB A100 GPUs for 10 epochs, completing the process in approximately 3 hours. The evaluation metrics follow previous works, using success rates based on predefined conditions for both open-loop and closed-loop tasks.  More details about the dataset can be found in Appendix.

\begin{figure*}[t]
\begin{center}
   \includegraphics[width=0.99\textwidth]{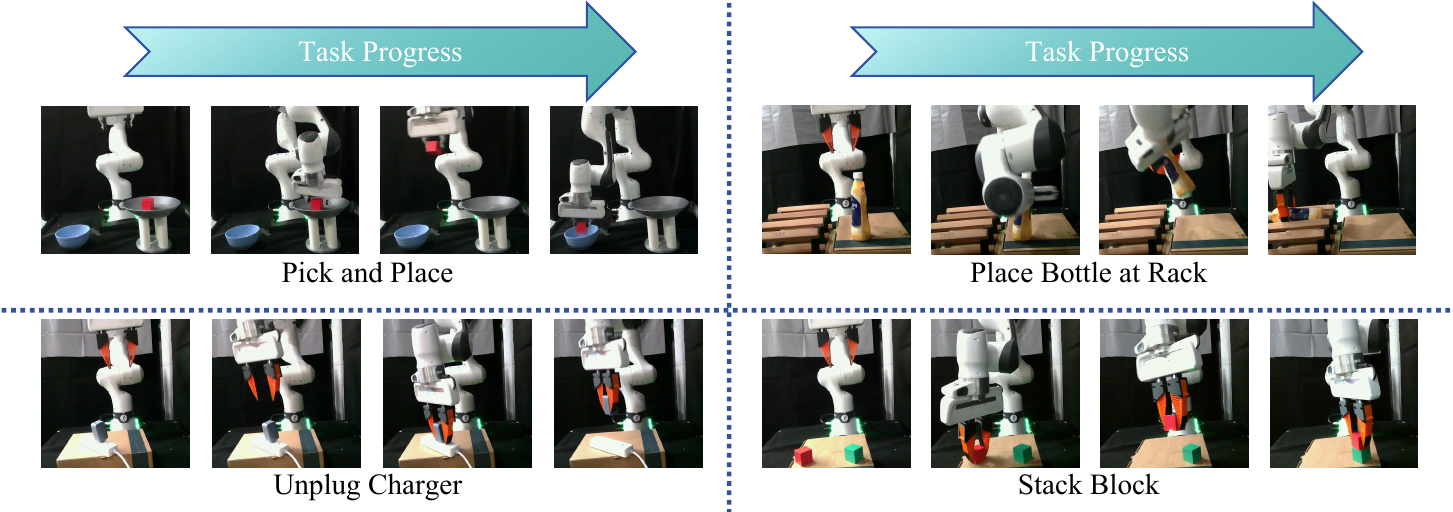}
\vspace{-0.1cm}
\caption{
\textbf{Task Execution Flow Illustrated with Keyframes. } 
We visualize the agent’s execution process for each task from the front perspective. The flow is from left to the right. 
}
   \label{fig:task_describe}
\end{center}
\vspace{-0.3cm}
\end{figure*}

\subsection{Simulation results}
\label{sec:MMS}

For the open-loop experiment, we compare our SC-VLA against four representative baselines: UMPNet~\cite{xu2022universal}, Flowbot3D~\cite{eisner2022flowbot3d}, ManipLLM~\cite{li2024manipllm}, and our utilized experts. 
Specifically, UMPNet~\cite{xu2022universal} inputs visual observations, such as RGB images and depth maps, of an articulated object to generate sequence of actions in SE(3) space.
Flowbot3D~\cite{eisner2022flowbot3d} takes point cloud as input, and select the predicted 3D flow with the maximum magnitude to interact with objects.
ManipLLM~\cite{li2024manipllm} uses chain-of-thought reasoning to enable the model to precisely generate an initial contact end-effector pose.
To ensure a fair comparison, all methods use the same train/test split and end-effector settings. Reproduction details of the baselines are provided in the Appendix.

As shown in Table~\ref{tab:main}, using our combined fast and slow systems Ours(Fast+Slow) achieves an impressive 87\% accuracy on seen tasks and 68\% accuracy on unseen tasks, significantly outperforming the other methods.
Specifically, compared to the previous SOTA ManipLLM, Ours(Fast+Slow) achieves improvements of 30\% and 21\% in accuracy for seen and unseen tasks, respectively. These results demonstrate that SC-VLA can effectively correct failed actions through the slow system's Chain of Thought (CoT) reasoning.
Additionally, the experts we employed achieved only 36\% and 39\% accuracy on seen and unseen categories, respectively. While these experts can provide potential position and rotation estimates, simply combining their outputs does not yield an optimal 6-DoF pose. The lower accuracy compared to our method demonstrates that our advantages stem not solely from the use of experts but from the overall effectiveness of our approach.
When comparing Ours(CPL+Fast) and Ours(Fast), we find that Ours(CPL+Fast) achieves improvements of 13\% and 39\% in accuracy for seen and unseen tasks. This indicates that our method effectively enhances the model’s manipulation accuracy and adaptability to varying scene configurations.
The significant improvement observed in unseen tasks further demonstrates that our proposed method can effectively enhance the model's generalization capabilities.

As for the close-loop experiment in RLBench~\cite{james2020rlbench}, Ours(Fast) achieves an average success rate of 63\% on the following five tasks ``take USB out of computer," ``close fridge," ``close box," ``toilet seat down," and ``unplug charger." In contrast, Ours (Fast+Slow) achieves a success rate of 91\%
This demonstrates that our SC-VLA effectively handles sequential action prediction and correction, further highlighting its robustness and effectiveness.
Due to space limitations, a comprehensive description of the closed-loop experiment and its settings is provided in the Appendix. 

\begin{figure*}[t]
\begin{center}
   \includegraphics[width=0.98\textwidth]{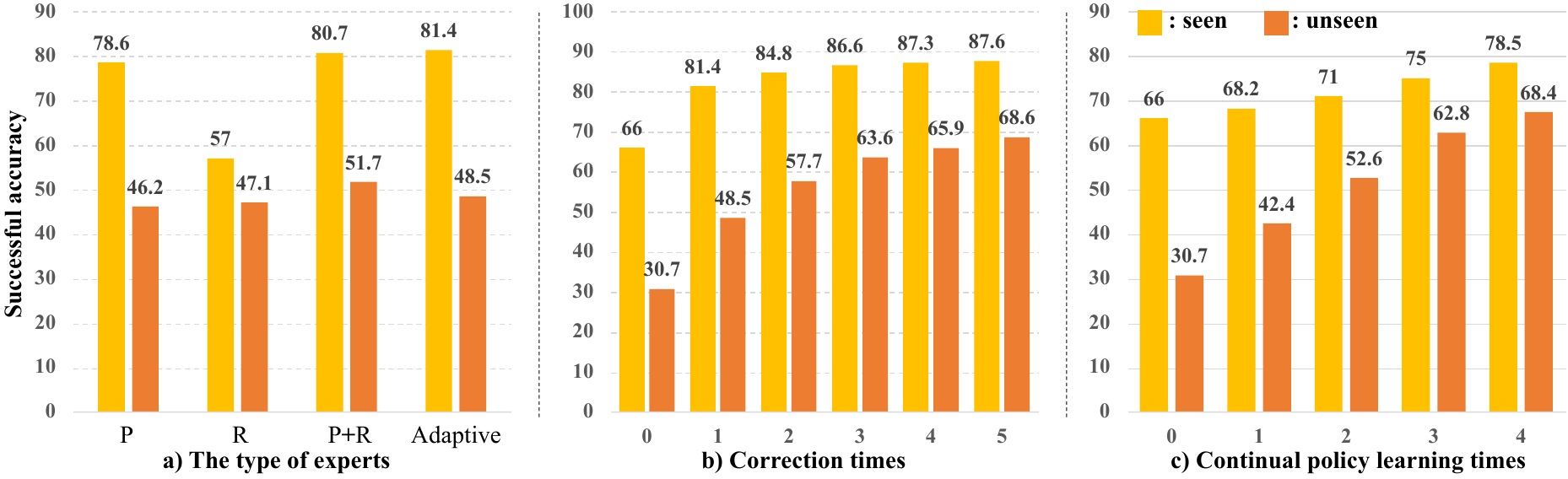}
\vspace{-0.2cm}
\caption{
Ablation study for each method in our SC-VLA framework. a) examines the impact of various expert feedback prompts on slow system reasoning. b) investigates the effects of different correction times on pose correction accuracy. c) analyzes the influence of continuous policy learning iterations on the model's prediction accuracy.
}
   \label{fig:abl}
\end{center}
\vspace{-0.35cm}
\end{figure*}
\begin{table}[t]
\centering
\caption{\textbf{Real-world experiments.} We test each method 10 times across diverse object poses.}
\vspace{-0.1cm}
\setlength\tabcolsep{5pt}
\small
\begin{tabular}{c|cccc|c}
\hline
 & Pick  & Place & Stack & Unplug \\
Method & place & bottle & block & charger & Mean\\
\midrule
ManipLLM  &0.40 & 0.20 & 0.10 & 0.20& 0.23 \\
DP3    &0.40 &0.50 &0.30 & 0.50 & 0.43 \\
Ours(Fast+Slow)  & \textbf{0.50} & \textbf{0.60} & \textbf{0.50} & \textbf{0.70} & \textbf{0.58} \\
\hline
\end{tabular}
\vspace{-0.2cm}
\label{aptab:real}
\end{table}

\subsection{Real-world results}
\label{sec:RWE}

\textbf{Experiment setting.}
In Figure~\ref{fig:task_describe}, we present real-world experiments using the SC-VLA model with a Franka Research 3 (FR3) robot and a 3D-printed UMI gripper~\cite{chi2024universal} across four manipulation tasks, including \textit{pick and place}, \textit{unplug charger}, \textit{place bottle at rack}, and \textit{stack blocks}.
We use a RealSense L515 camera to capture real-world visual observations from the front view. 
For each task, we collect 30 demonstrations with randomized object poses via teleoperation. Key frames are then extracted to construct the training set. We train our method for 10 epochs, loading simulation-pretrained model parameters. Since fine-tuning on real-world data is primarily aimed at addressing the sim-to-real gap, we use the simple fast system prompt for training.
For evaluation, we use the model from the final training epoch and test it on 10 trials with diverse object poses for a robust evaluation.

\textbf{Quantitative results.} 
As shown in Table~\ref{aptab:real}, we evaluate the effectiveness of SC-VLA by comparing it with two methods: (1) ManipLLM~\cite{li2023manipllm}, where we load its simulation-pretrained parameters for a fair comparison, and (2) 3D Diffusion Policy (DP3)~\cite{ze20243d}, a state-of-the-art (SOTA) point cloud-based method that employs diffusion-based generation for action prediction.
Compared to ManipLLM, SC-VLA outperforms it in multiple tasks. Since both methods adopt the same backbone, this comparison directly reflects our method's Chain-of-Thought reasoning ability to recognize failures, request feedback from experts, and correct action poses.
Notably, in the \textit{stack block} task, which demands high precision in pose prediction, SC-VLA achieves a 50\% success rate, representing a 40\% improvement over ManipLLM.
Furthermore, compared to DP3, our SC-VLA achieves superior performance across various tasks, resulting in a 15\% performance improvement. This further demonstrates SC-VLA’s ability to predict accurate contact points and 3D directions in real-world scenarios.

\textbf{Qualitative results.}
As shown in Figure~\ref{fig:task_describe}, we visualize the key frames of manipulation process for four real-world tasks. Our method accurately predicts the 7-DoF end-effector poses, enabling successful task completion along the predicted trajectories.
For instance, in the \textit{place bottle at rack} task, our method first secures a precise grasp near the bottle's neck, then rotates the gripper to the optimal placement angle, and finally determines the perfect timing to open the gripper for placement.
More real-world demonstrations are shown in the supplementary video.

\label{sec:Experiment}

\subsection{Ablation study}
\label{sec:AS}
To elucidate the contribution and effectiveness of individual methods within our SC-VLA, we conduct extensive ablation studies in the SAPIEN simulator.

\textbf{The impact of expert type.}
First, we explore the influence of different types of expert feedback prompts on the success rate of corrections performed by the slow system. As shown in Figure~\ref{fig:abl} a), ``P" and ``R" represent utilizing our designed position and rotation feedback, respectively. ``P + R" represents the utilization of combined expert feedback, while ``Adaptive" (ours) refers to adaptively seeking expert feedback based on the cause of failure. For all experiments, we input the prompt feedback into our SC-VLA, allowing it to re-predict end-effector poses. We find that any type of expert prompt can improve unseen manipulation accuracy, demonstrating the importance of correction for novel object manipulation. Additionally, we observe that ours (``Adaptive") achieves comparable results to ``P + R", but with lower expert intervention costs. The results confirm that detecting failure cases and adaptively seeking expert feedback is essential for stable manipulation. The accuracy of failure detection is shown in the Appendix.

\textbf{The impact of correction times.}
In the slow system, our experts can return multiple prompts simultaneously; for example, the position expert can provide $n$ potential contact points. Therefore, we explore the impact of different correction times on pose correction accuracy. As shown in Figure~\ref{fig:abl} b), the x-axis represents the number of expert prompts used for correction. Note that we count a manipulation as successful if it succeeds even once during multiple corrected actions. We find that with just one correction, SC-VLA achieves an improvement of 15.4\% on seen tasks and 17.8\% on unseen tasks compared to our fast system. The results demonstrate the effectiveness of our designed CoT correction paradigm. Furthermore, once the number of corrections reaches three or more, the slow system achieves robust manipulation accuracy.

\textbf{The impact of continuous policy learning times}

After correction by the slow system, SC-VLA can output the correct pose and rotation. We save the successfully corrected samples and use them for continuous policy learning. The varying update iterations will impact the fine-tuning time and the manipulation accuracy achieved after fine-tuning.
As shown in Figure~\ref{fig:abl} c), all manipulation results are evaluated using the fast system's inference manner after the continual learning process.
The x-axis represents the training iteration for each corrected sample in continual policy learning. Both seen and unseen tasks improve significantly after multiple fine-tuning rounds, validating the effectiveness of our method in transferring knowledge from the slow system to the fast system. To rule out overfitting on the test set, we also evaluate the model on the test B set after continuous learning (shown in the Appendix).

\vspace{-0.1em}
\section{Conclusion and limitation}

Drawing on Daniel Kahneman's concept of a ``fast system" and ``slow system," we propose a self-corrected VLA that emulates human-like thinking for stable manipulation. SC-VLA combines a fast system for direct end-effector pose prediction and a slow system for CoT reflecting on and correcting failed actions. During inference, the fast system interacts with the physical world, while the slow system activates only when a failure occurs. We also introduce continuous policy learning, allowing SC-VLA to learn from corrected actions and continuously improve the fast system's accuracy.
Experimental results demonstrate that SC-VLA significantly enhances manipulation performance in both simulation and real-world scenarios.

A limitation of SC-VLA is that our slow system chain-of-thought reasoning leads to additional computational cost, which can be overcomed by integrating more efficient LLMs~\cite{wen2024tinyvla, liu2024robomamba, zhou2024tinyllava}.

\label{sec:conclusion}

\vspace{-0.1em}
\section{Acknowledge}
This work was supported by the National Natural Science Foundation of China (62476011).
\label{sec:acknowledge}

{
    \small
    \bibliographystyle{ieeenat_fullname}
    \bibliography{citations}
}

\clearpage

\setcounter{page}{1}
\appendix

\begin{center}
    \Large
    \vspace{0.5em} Appendix \\
    \vspace{1.0em}
\end{center}

Due to space limitations, we provide additional details on the dataset, baseline reproduction, SC-VLA failure detection evaluation, and supplementary experiments. Below, we outline the contents of our appendix.

\noindent\hspace*{0.1em} \textbf{A. Additional simulation experiment setting}   

\noindent\hspace*{0.1em} \textbf{B. Dataset \& baseline reproduction details} 

\noindent\hspace*{0.1em} \textbf{C. Failure detection results}

\noindent\hspace*{0.1em} \textbf{D. Additional ablation study}

\noindent\hspace*{0.1em} \textbf{E. Experimental results on Test B set}

\noindent\hspace*{0.1em} \textbf{F. Details of Close-loop Experiments}

\noindent\hspace*{0.1em} \textbf{G. Additional related work}

\noindent\hspace*{0.1em} \textbf{H. Model architecture of SC-VLA}

\noindent\hspace*{0.15em} \textbf{I. Advantages of fast \& slow systems}

\section{Additional simulation experiment setting}
\label{apsec:ASES}
\subsection{Data Collection}
We use the simulator to collect training data under open-loop and closed-loop control.
For the open-loop experiment, we follow the data collection process of previous works~\citep{mo2021where2act, li2023manipllm}, adopting the SAPIEN engine~\citep{Xiang_2020_SAPIEN} to set up an interactive simulation environment with articulated objects from PartNet-Mobility~\citep{mo2019partnet}. The Franka Panda Robot, equipped with a suction gripper, serves as the robotic actuator. During data collection, we randomly select a contact point \textbf{p} on the movable part and orient the end-effector's z-axis opposite to its normal vector, with a random y-axis direction to interact with the object. Successful operations are categorized as successful samples and integrated into the dataset.
Our training dataset comprises 12k successful manipulation samples across 20 categories. 
Meanwhile, we collect 15k failure samples and 60k corresponding correction prompts, covering position, rotation, and combined errors.
For evaluation, we generate 1k examples for the test set, comprising 20 training (seen) and 10 testing (unseen) categories. The unseen categories are used to evaluate the generalization capability of SC-VLA.
The additional description of the dataset and variation can be found in Appendix~\ref{sec:dataset} and ~\ref{apsec:sdv}.
For the closed-loop experiment, we select five tasks from RLBench~\citep{james2020rlbench}, including ``take USB out of computer," ``close fridge," ``close box," ``toilet seat down," and ``unplug charger." We utilize a single-view input (the front view from a third-person perspective) and follow to the key frame selection manner outlined in previous work~\cite{shridhar2023perceiver}.

\subsection{Implementation Details}

Our SC-VLA loads the pre-trained parameters of LLaMA-Adapter-v2\citep{zhang2023llama}, which integrates a pre-trained CLIP\citep{radford2021learning} as the visual encoder and a 7B LLaMA\citep{touvron2023llama} model as the language model. The multi-modal projection module is built with 32 transformer layers, each configured with a hidden dimension of 4096—carefully chosen as a multiple of 256—to ensure efficient scaling. Each layer also features 32 attention heads and employs a layer normalization epsilon of 1e-05 for enhanced stability, while the vocabulary size is set to -1 to enable flexible token handling. Throughout the fine-tuning phase, we utilize the Adam optimizer with ($(\beta_{1}, \beta_{2}) = (0.9, 0.999)$ and an initial learning rate of 1e-4, accompanied by a warm-up period of one epoch. We fine-tuned our model on four 80G A100 GPUs for 10 epochs.

\subsection{Evaluation Metric}
For the open-loop experiment, we follow the metrics from previous works~\citep{li2023manipllm}, using the manipulation success rate. Specifically, the object starts in its initial state, and the goal is to actuate the joint to its target state. We use the predicted contact point and rotation to interact with objects and complete the task.
Similar to \citep{mo2021where2act}, the trajectory for each task is predefined, i.e., moving backward along the z-axis of the end-effector pose. An action is considered successful if the joint state difference before and after interaction exceeds a threshold of 0.1 meters. For the closed-loop experiment, we follow the evaluation metrics from previous works\citep{james2020rlbench}, assessing task success rate based on predefined success conditions.

\section{Dataset \& baseline reproduction details}
\label{apsec:DBRD}
\subsection{Sapien Dataset Details}
\label{sec:dataset}
As shown in Table~\ref{tab:icon}, our training dataset comprises 12k manipulation scenarios, encompassing 20 distinct object categories, specifically including: Safe (\textit{S}), Door (\textit{D}), Display (\textit{DS}), Refrigerator (\textit{RF}), Laptop (\textit{LT}), Lighter (\textit{LI}), Microwave (\textit{MW}), Mouse (\textit{MO}), Box (\textit{BX}), Trash Can (\textit{TC}), Kitchen Pot (\textit{KP}), Suitcase (\textit{SU}), Pliers (\textit{PL}), Storage Furniture (\textit{SF}), Remote (\textit{RM}), Bottle (\textit{B}), Folding Chair (\textit{FD}), Toaster (\textit{TS}), Lamp (\textit{L}), and Dispenser (\textit{DP}).
Following Partnet~\citep{mo2019partnet}, different tasks are designed for each category. For instance, opening the door or control panel of a refrigerator, opening the cap of a bottle, and rotating the lid of a box. The detailed task design can be found at: \href{https://sapien.ucsd.edu/browse}{https://sapien.ucsd.edu/browse}

In performance evaluation experiments, we utilize two primary test datasets: Test set and Test B set. Both datasets consist of 1081 manipulation scenarios and include 30 object categories, as detailed in Table~\ref{tab:icon}. 
We collect the test B set where only the relative positions between the robot and the object are altered compared to the test set. 
This test B set aims to validate the effectiveness of our continuous policy learning, ensuring it does not overfit on the test set.
Among these, 20 categories are present in the training set (seen), while 10 categories are not included in the training set (unseen), which are: Toilet (\textit{TL}), Scissors (\textit{SC}), Table (\textit{T}), Stapler (\textit{ST}), Kettle (\textit{K}), USB (\textit{U}), Switch (\textit{SW}), Washing Machine (\textit{WM}), Faucet (\textit{FC}), and Phone (\textit{PH}). This setup aims to thoroughly assess the model's generalization capabilities. 

\begin{table*}[htbp]
    \centering
    \small
    \renewcommand{\arraystretch}{1.2}
    \caption{Representation of each category icon.}
    \label{tab:icon}
    \setlength{\tabcolsep}{0.05pt}
    \resizebox{\textwidth}{!}{
    \begin{tabular}{>{\centering\arraybackslash}m{1.5cm} >{\centering\arraybackslash}m{1.5cm} >{\centering\arraybackslash}m{1.5cm} >{\centering\arraybackslash}m{1.5cm} >{\centering\arraybackslash}m{1.5cm} >{\centering\arraybackslash}m{1.5cm} >{\centering\arraybackslash}m{1.5cm} >{\centering\arraybackslash}m{1.5cm} >{\centering\arraybackslash}m{1.5cm} >{\centering\arraybackslash}m{1.5cm}}
        \hline
        \includegraphics[width=0.50\linewidth]{./images/icon/safe.png} &
        \includegraphics[width=0.50\linewidth]{./images/icon/door.png} &
        \includegraphics[width=0.50\linewidth]{./images/icon/display.png} &
        \includegraphics[width=0.50\linewidth]{./images/icon/refrigerator.png} &
        \includegraphics[width=0.50\linewidth]{./images/icon/laptop.png} &
        \includegraphics[width=0.50\linewidth]{./images/icon/lighter.png} &
        \includegraphics[width=0.50\linewidth]{./images/icon/microwave.png} &
        \includegraphics[width=0.50\linewidth]{./images/icon/mouse.png} &
        \includegraphics[width=0.50\linewidth]{./images/icon/box.png} &
        \includegraphics[width=0.50\linewidth]{./images/icon/trash_can.png} \\
        \hline
        \textbf{\textit{Safe}} & \textbf{\textit{Door}} & \textbf{\textit{Display}} & \textbf{\textit{Refrigerator}} & \textbf{\textit{Laptop}} & \textbf{\textit{Lighter}} & \textbf{\textit{Microwave}} & \textbf{\textit{Mouse}} & \textbf{\textit{Box}} & \textbf{\textit{Trashcan}} \\
        \hline\hline
        \includegraphics[width=0.50\linewidth]{./images/icon/kitchen_pot.png} &
        \includegraphics[width=0.50\linewidth]{./images/icon/suitcase.png} &
        \includegraphics[width=0.50\linewidth]{./images/icon/pliers.png} &
        \includegraphics[width=0.50\linewidth]{./images/icon/storage_furniture.png} &
        \includegraphics[width=0.50\linewidth]{./images/icon/remote.png} &
        \includegraphics[width=0.50\linewidth]{./images/icon/bottle.png} &
        \includegraphics[width=0.50\linewidth]{./images/icon/folding_chair.png} &
        \includegraphics[width=0.50\linewidth]{./images/icon/toaster.png} &
        \includegraphics[width=0.50\linewidth]{./images/icon/lamp.png} &
        \includegraphics[width=0.50\linewidth]{./images/icon/dispenser.png} \\
        \hline
        \textbf{\textit{Kitchen Pot}} & \textbf{\textit{Suitcase}} & \textbf{\textit{Pliers}} & \textbf{\textit{Storage Furniture}} & \textbf{\textit{Remote}} & \textbf{\textit{Bottle}} & \textbf{\textit{Folding Chair}} & \textbf{\textit{Toaster}} & \textbf{\textit{Lamp}} & \textbf{\textit{Dispenser}} \\
        \hline\hline
        \includegraphics[width=0.50\linewidth]{./images/icon/toilet.png} &
        \includegraphics[width=0.50\linewidth]{./images/icon/scissors.png} &
        \includegraphics[width=0.50\linewidth]{./images/icon/table.png} &
        \includegraphics[width=0.50\linewidth]{./images/icon/stapler.png} &
        \includegraphics[width=0.50\linewidth]{./images/icon/kettle.png} &
        \includegraphics[width=0.50\linewidth]{./images/icon/usb.png} &
        \includegraphics[width=0.50\linewidth]{./images/icon/switch.png} &
        \includegraphics[width=0.50\linewidth]{./images/icon/washing_machine.png} &
        \includegraphics[width=0.50\linewidth]{./images/icon/faucet.png} &
        \includegraphics[width=0.50\linewidth]{./images/icon/phone.png} \\
        \hline
        \textbf{\textit{Toilet}} & \textbf{\textit{Scissors}} & \textbf{\textit{Table}} & \textbf{\textit{Stapler}} & \textbf{\textit{Kettle}} & \textbf{\textit{USB}} & \textbf{\textit{Switch}} & \textbf{\textit{Washing Machine}} & \textbf{\textit{Faucet}} & \textbf{\textit{Phone}} \\
        \hline
    \end{tabular}
    }
    \vspace{-0.2cm}
\end{table*}

\subsection{Sapien Dataset variation}
\label{apsec:sdv}
Regarding the variation between training and testing data, we followed the data collection settings of where2act~\citep{mo2021where2act} and ManipLLM~\citep{li2023manipllm}. The specific variations can be divided into two aspects: 1) Asset Variation and 2) State Variation.

1) Asset Variation: We use 20 categories from PartNet~\citep{mo2019partnet} for seen objects and reserve the remaining 10 categories for unseen objects to analyze if RoboMamba can generalize to novel categories. Specifically, we further divide the seen objects into 1037 training shapes and 489 testing shapes, using only the training shapes to construct the training data. Thus, the shapes of the seen objects encountered during training and testing are different. For unseen categories, there are a total of 274 shapes, which are used exclusively in the testing data.

2) State Variation: We observe the object in the scene from an RGB-D camera with known intrinsics, mounted 4.5-5.5 units away from the object, facing its center. The camera is located at the upper hemisphere of the object with a random azimuth between [-45, 45] and a random altitude between [30, 60]. Since the tasks involve 'pulling,' we also initialize the starting pose for each articulated part randomly between its rest joint state (fully closed) and any position up to half of its joint state (half-opened). These state settings are utilized for both training and testing data, aiming to boost the model's generalization ability.

\subsection{Baseline Reproduction}
\label{apsec:BR}

\textbf{UMPNet~\citep{xu2022universal}}: 
It inputs visual observations, such as RGB images and depth maps, of an articulated object to generate a sequence of actions in SE(3) space.
It identifies the correct position on the object to interact with (\textit{e.g.}., interacting with the lid rather than the base) and determines the appropriate action direction (\textit{e.g.}, pulling up instead of pushing down) for interaction.

\noindent\textbf{Flowbot3D~\citep{eisner2022flowbot3d}}:
Flowbot3D begins by observing the initial configuration of the object of interest in the form of point cloud data. 
It is then post-processed and inputted into the model, which predicts 3D flow vectors for each point. 
It selects the point with the maximum flow vector magnitude and uses motion planning to make contact with that point via suction based on the selected flow.

\noindent\textbf{ManipLLM~\citep{li2023manipllm}}: 
ManipLLM uses chain-of-thought reasoning to enable the model to precisely generate an initial contact end-effector pose, including the pixel coordinates, gripper upward direction, and gripper forward direction. 
It then employs an active impedance adaptation policy that adjusts the direction based on force feedback to ensure a smooth movement trajectory.

\section{Failure detection results}
\label{apsec:APFD}
In this paper, we introduce a self-corrected (SC) MLLM that mimics a human-like thinking paradigm, including two reasoning modes: fast system and slow system. SC-VLA possesses both the fast system's capability to directly predict end-effector poses and the slow system's ability to reflect on and correct failure actions.
In the slow system chain of thought reasoning, one important intermediate process is failure case detection.
The failure detection results are shown in Table \ref{tab:failure_detection}. The experiments are evaluated on a self-collected dataset, which contains 1K failure test samples on 20 seen categories of objects with three failure causes (i.e., position, rotation, or combined error).

\begin{table*}[htbp]
    \centering
    \small
    \setlength{\tabcolsep}{2pt}
    \renewcommand{\arraystretch}{1.2}
    \caption{Failure detection accuracy is evaluate on collected failure manipulation samples across 20 seen categories with three failure causes: position, rotation, or both.}
    \label{tab:failure_detection}

    \resizebox{\textwidth}{!}{
    \begin{tabular}{c|*{20}{c}c}
        \hline
        \multirow{2}{*}{\textbf{\raisebox{-1.0\height}{Failure Causes}}} & \multicolumn{21}{c}{\textbf{Object Categories}} \\
        \cline{2-22}
        
        & \includegraphics[width=0.05\linewidth]{./images/icon/microwave.png}
        & \includegraphics[width=0.05\linewidth]{./images/icon/mouse.png}
        & \includegraphics[width=0.05\linewidth]{./images/icon/phone.png}
        & \includegraphics[width=0.05\linewidth]{./images/icon/pliers.png}
        & \includegraphics[width=0.05\linewidth]{./images/icon/refrigerator.png}
        & \includegraphics[width=0.05\linewidth]{./images/icon/remote.png}
        & \includegraphics[width=0.05\linewidth]{./images/icon/safe.png}
        & \includegraphics[width=0.05\linewidth]{./images/icon/scissors.png}
        & \includegraphics[width=0.05\linewidth]{./images/icon/bottle.png}
        & \includegraphics[width=0.05\linewidth]{./images/icon/box.png}
        & \includegraphics[width=0.05\linewidth]{./images/icon/dispenser.png}
        & \includegraphics[width=0.05\linewidth]{./images/icon/display.png}
        & \includegraphics[width=0.05\linewidth]{./images/icon/door.png}
        & \includegraphics[width=0.05\linewidth]{./images/icon/faucet.png}
        & \includegraphics[width=0.05\linewidth]{./images/icon/folding_chair.png}
        & \includegraphics[width=0.05\linewidth]{./images/icon/kettle.png}
        & \includegraphics[width=0.05\linewidth]{./images/icon/kitchen_pot.png}
        & \includegraphics[width=0.05\linewidth]{./images/icon/lamp.png}
        & \includegraphics[width=0.05\linewidth]{./images/icon/laptop.png}
        & \includegraphics[width=0.05\linewidth]{./images/icon/lighter.png}
        & \textbf{\raisebox{0.5\height}{AVG}} \\
        \hline\hline

        \textit{Rotation} & 0.95 & 0.84 & 0.88 & 0.80 & 0.81 & 0.96 & 0.88 & 0.82 & 0.94 & 0.78 & 0.93 & 0.73 & 0.83 & 0.89 & 0.93 & 0.94 & 1.00 & 0.82 & 0.95 & 0.95 & \textbf{0.89} \\
        
        \textit{Position} & 0.95 & 1.00 & 1.00 & 1.00 & 1.00 & 1.00 & 0.95 & 1.00 & 0.94 & 1.00 & 0.95 & 1.00 & 1.00 & 0.93 & 1.00 & 0.91 & 1.00 & 1.00 & 1.00 & 0.95 & \textbf{0.98} \\
        
        \textit{Position \& Rotation} & 0.71 & 0.74 & 0.92 & 0.88 & 0.83 & 1.00 & 0.71 & 0.91 & 0.91 & 0.95 & 0.83 & 0.82 & 0.82 & 0.82 & 0.92 & 0.84 & 1.00 & 0.84 & 1.00 & 0.61 & \textbf{0.84} \\
        \hline
    \end{tabular}
    }
    \vspace{-0.1cm}
\end{table*}

SC-VLA achieves an impressive average accuracy of \textbf{0.89} for rotation-related failures. This high accuracy underscores the model's capability to precisely identify and diagnose issues related to the robot's manipulation direction. Notably, categories such as \textit{Remote}, \textit{Lamp}, and \textit{Laptop} exhibit exceptional accuracy rates, reflecting the model's robustness across varied object geometries and complexities.

For position-related failures, SC-VLA achieves an outstanding average accuracy of \textbf{0.98}. This near-perfect detection rate emphasizes the reliability of our SC-VLA in pinpointing positional inaccuracies during manipulation actions. Categories including \textit{Mouse}, \textit{Phone }, \textit{Pliers}, and \textit{Refrigerator} all attain a flawless accuracy of 1.00, demonstrating the SC-VLA's precision in handling positional errors.

When addressing failures caused by both position and rotation, the model maintains a commendable average accuracy of \textbf{0.84}. Despite the increased complexity of these cases, our method effectively diagnoses combined failures, ensuring that subsequent correction experts are accurately requested. Categories such as \textit{Remote}, \textit{Kettle}, and \textit{Laptop} once again exhibit high accuracy rates, reaffirming the model's adaptability and precision.

These results highlight the effectiveness of our SC-VLA in failure case detection and reflection. Accurately identifying the failure cause can significantly enhance the stability of our slow system reasoning, thereby improving failure correction. 

\section{Additional ablation study}
\label{sec:ABS}
To further validate the robustness of our proposed SC-VLA, we conduct supplementary ablation studies. In this section, we assess the efficacy of our approach in the context of continual policy learning. Specifically, we integrate various continual learning methods into our self-correction process to update the model and evaluate the performance of the updated model. The test set used for this evaluation, is consistent with the experimental setup described in the main text. It comprises 30 categories of manipulation targets, totaling 1081 manipulation scenarios. The results of these evaluations are presented in Table \ref{tab:continual_learning}, demonstrating the performance improvements achieved through the integration of continual learning methods in our method. After each update, the model undergoes a new round of manipulation tests, and successful manipulation instances are collected. These successful examples were then incorporated as new data for continual learning, further updating the model. All test results are obtained after the base model is updated four times continually, with performance measured and recorded accordingly.

\begin{table*}[htbp]
    \centering
    \small
    \setlength{\tabcolsep}{2pt} 
    \renewcommand{\arraystretch}{1.2}
    \caption{
    Comparisons of our proposed method against other continual learning methods. The table presents the performance of different methods across various seen and unseen categories. "Ours(CPL+Fast)" represents our method's results for continually learned policies without slow system reasoning.}
    \label{tab:continual_learning}
    \resizebox{\textwidth}{!}{
    \begin{tabular}{c|cccccccccccccccc}
        \hline
        \multirow{2}{*}{\textbf{\raisebox{-1.0\height}{Method}}} & \multicolumn{16}{c}{\textbf{Seen Categories}} \\
        \cline{2-17}
        & \includegraphics[width=0.050\linewidth]{./images/icon/safe.png}
        &\includegraphics[width=0.050\linewidth]{./images/icon/door.png}
        &\includegraphics[width=0.050\linewidth]{./images/icon/display.png}
        &\includegraphics[width=0.050\linewidth]{./images/icon/refrigerator.png}
        &\includegraphics[width=0.050\linewidth]{./images/icon/laptop.png}
        &\includegraphics[width=0.050\linewidth]{./images/icon/lighter.png}
        &\includegraphics[width=0.050\linewidth]{./images/icon/microwave.png}
        &\includegraphics[width=0.050\linewidth]{./images/icon/mouse.png}
        &\includegraphics[width=0.050\linewidth]{./images/icon/box.png}
        &\includegraphics[width=0.050\linewidth]{./images/icon/trash_can.png}
        &\includegraphics[width=0.050\linewidth]{./images/icon/kitchen_pot.png}
        &\includegraphics[width=0.050\linewidth]{./images/icon/suitcase.png}
        &\includegraphics[width=0.050\linewidth]{./images/icon/pliers.png}
        &\includegraphics[width=0.050\linewidth]{./images/icon/storage_furniture.png}
        &\includegraphics[width=0.050\linewidth]{./images/icon/remote.png}
        &\includegraphics[width=0.050\linewidth]{./images/icon/bottle.png}\\
        \hline\hline 
        EWC~\cite{kirkpatrick2017overcoming} & 0.71 & 0.80 & 0.33 & 0.72 & 0.71 & 0.40 & 0.78 & 0.96 & 0.56 & 0.71 & 0.85 & 0.80 & 0.38 & 0.76 & 0.15 & 0.90 \\
        LwF~\cite{li2017learning} & 0.88 & 0.73 & 0.41 & 0.81 & 0.90 & 0.40 & 0.83 & 0.88 & 0.37 & 0.82 & 0.90 & 0.84 & 0.51 & 0.88 & 0.30 & 0.93 \\
        Experience Replay~\cite{rolnick2019experience} & 0.83 & 0.73 & 0.50 & 0.84 & 0.85 & 0.26 & 0.72 & 0.88 & 0.56 & 0.75 & 0.90 & 0.79 & 0.42 & 0.83 & 0.23 & 0.90 \\
        \hline
        Ours(CPL+Fast) & 0.90 & 0.75 & 0.58 & 0.87 & 0.95 & 0.46 & 0.89 & 0.92 & 0.50 & 0.78 & 0.90 & 0.85 & 0.63 & 0.90 & 0.38 & 0.90 \\
        \hline
    \end{tabular}
    }
    \vspace{0.5em}
    \resizebox{\textwidth}{!}{
    \begin{tabular}{c|ccccc|ccccccccccc}
        \hline
        \multirow{2}{*}{\textbf{\raisebox{-1.0\height}{Method}}} & \multicolumn{5}{c|}{\textbf{Seen Categories}} & \multicolumn{11}{c}{\textbf{Unseen Categories}}\\
        \cline{2-17}
        & \includegraphics[width=0.050\linewidth]{./images/icon/folding_chair.png}
        &\includegraphics[width=0.050\linewidth]{./images/icon/toaster.png}
        &\includegraphics[width=0.050\linewidth]{./images/icon/lamp.png}
        &\includegraphics[width=0.050\linewidth]{./images/icon/dispenser.png}
        &\multicolumn{1}{c|}{{\textbf {\raisebox{0.5\height}{AVG}}} }
        &\includegraphics[width=0.050\linewidth]{./images/icon/toilet.png}
        &\includegraphics[width=0.050\linewidth]{./images/icon/scissors.png}
        &\includegraphics[width=0.050\linewidth]{./images/icon/table.png}
        &\includegraphics[width=0.050\linewidth]{./images/icon/stapler.png}
        &\includegraphics[width=0.050\linewidth]{./images/icon/kettle.png}
        &\includegraphics[width=0.050\linewidth]{./images/icon/usb.png}
        &\includegraphics[width=0.050\linewidth]{./images/icon/switch.png}
        &\includegraphics[width=0.050\linewidth]{./images/icon/washing_machine.png}
        &\includegraphics[width=0.050\linewidth]{./images/icon/faucet.png}
        &\includegraphics[width=0.050\linewidth]{./images/icon/phone.png}
        &{\textbf {\raisebox{0.5\height}{AVG}}} \\
        \hline\hline 
        EWC~\cite{kirkpatrick2017overcoming} & 0.20 & 0.68 & 0.53 & 0.90 & 0.68    & 0.00 & 0.29 & 0.50 & 0.33 & 0.35 & 0.52 & 0.47 & 0.56 & 0.25 & 0.87 & 0.42 \\
        LwF~\cite{li2017learning} & 0.40 & 0.73 & 0.67 & 0.80 & 0.75 & 0.27 & 0.53 & 0.67 & 0.83 & 0.92 & 0.64 & 0.65 & 0.78 & 0.47 & 0.93 & 0.62 \\
        Experience Replay~\cite{rolnick2019experience} & 0.00 & 0.70 & 0.61 & 0.80 & 0.71 & 0.00 & 0.40 & 0.61 & 0.66 & 0.57 & 0.58 & 0.56 & 0.68 & 0.34 & 0.93 & 0.52 \\
        \hline
Ours(CPL+Fast) & 0.60 & 0.71 & 0.74 & 0.90 & 0.79 & 0.27 & 0.61 & 0.71 & 0.50 & 0.92 & 0.69 & 0.69 & 0.80 & 0.70 & 0.81 & 0.69 \\
        \hline
    \end{tabular}
    }
    \vspace{-0.1cm}
    \vspace{-0.1cm}
\end{table*}

As shown in Table \ref{tab:continual_learning}, Ours(CPL+Fast) seamlessly integrates Exponential Moving Average (EMA)  to continually fine-tune the injected adapters, showing significant performance improvements across multiple evaluation metrics. Our method surpasses Elastic Weight Consolidation (EWC), Learning without Forgetting (LwF), and Experience Replay in both seen and unseen categories. Specifically, in the seen categories, our method achieves an impressive average successful rate of 0.79, compared to 0.68 for EWC, 0.75 for LwF, and 0.71 for Experience Replay. This superior performance indicates that our approach not only retains knowledge from previous tasks but also effectively assimilates new information, thereby addressing the prevalent problem of catastrophic forgetting.

In the unseen categories, our method consistently maintains a high performance with an average score of 0.69, compared to 0.42 for EWC, 0.62 for LwF, and 0.52 for Experience Replay. This consistent performance across both seen and unseen categories underscores the robustness of our continual learning strategy. Notably, our method excels in categories such as \textit{Laptop (LT)}, \textit{Phone (PH)}, and \textit{Refrigerator (RF)}, demonstrating its effectiveness to generalize across a diverse array of objects and scenarios.

Furthermore, our approach exhibits marked improvement in categories with lower baseline performances. For instance, in the \textit{Dispenser (DP)} category, our method achieves a perfect score of 0.95, highlighting its capacity to manage challenging tasks with remarkable efficacy. Similarly, in the \textit{Lamp (L)} and \textit{Kettle (K)} categories, our method significantly outperforms other continual learning methods, achieving scores of 0.74 and 0.92, respectively.
Overall, the experimental results validate the effectiveness of our proposed continuous policy learning. By incorporating the Exponential Moving Average scheme in the fine-tuning process, our approach ensures robust performance improvements while mitigating the risks of catastrophic forgetting. This enables our model to efficiently transfer successfully corrected samples from the slow system to the fast system.

\section{Experimental results on Test B set}
\label{sec:TestB}

In this section, we evaluate our proposed method on the Test B set following the continuous learning process, aiming to demonstrate that the observed improvements are not simply due to overfitting on the test set.
Similar to the Test set, the Test B set consists of 1,081 manipulation scenarios involving 30 different objects. To simulate real-world applications, we adjusted the relative positions between the robot and the objects in the Test B set compared to the Test set.
It is important to note that the Test B set is not used for any fine-tuning. The model's performance on the Test B set is recorded after each iteration of continual policy learning on the Test set.

\begin{table*}[htbp]
    \centering
    \small
    \setlength{\tabcolsep}{2pt}
    \renewcommand{\arraystretch}{1.2}
    \caption{
    \textbf{Performance on Test B set.} The success rate is recorded after each iteration of continual policy learning on the Test set.
    }
    \label{tab:testB}
    \resizebox{\textwidth}{!}{
    \begin{tabular}{c|cccccccccccccccc}
        \hline
        \multirow{2}{*}{\textbf{\raisebox{-1.0\height}{Method}}} & \multicolumn{16}{c}{\textbf{Seen Categories}} \\
        \cline{2-17}
        &\includegraphics[width=0.050\linewidth]{./images/icon/safe.png}
        &\includegraphics[width=0.050\linewidth]{./images/icon/door.png}
        &\includegraphics[width=0.050\linewidth]{./images/icon/display.png}
        &\includegraphics[width=0.050\linewidth]{./images/icon/refrigerator.png}
        &\includegraphics[width=0.050\linewidth]{./images/icon/laptop.png}
        &\includegraphics[width=0.050\linewidth]{./images/icon/lighter.png}
        &\includegraphics[width=0.050\linewidth]{./images/icon/microwave.png}
        &\includegraphics[width=0.050\linewidth]{./images/icon/mouse.png}
        &\includegraphics[width=0.050\linewidth]{./images/icon/box.png}
        &\includegraphics[width=0.050\linewidth]{./images/icon/trash_can.png}
        &\includegraphics[width=0.050\linewidth]{./images/icon/kitchen_pot.png}
        &\includegraphics[width=0.050\linewidth]{./images/icon/suitcase.png}
        &\includegraphics[width=0.050\linewidth]{./images/icon/pliers.png}
        &\includegraphics[width=0.050\linewidth]{./images/icon/storage_furniture.png}
        &\includegraphics[width=0.050\linewidth]{./images/icon/remote.png}
        &\includegraphics[width=0.050\linewidth]{./images/icon/bottle.png} \\
        \hline\hline 
        Ours-cl-turn1 & 0.69 & 0.70 & 0.25 & 0.81 & 0.71 & 0.33 & 0.83 & 0.88 & 0.62 & 0.60 & 1.00 & 0.74 & 0.31 & 0.71 & 0.07 & 0.93 \\
        Ours-cl-turn2 & 0.71 & 0.58 & 0.16 & 0.67 & 0.85 & 0.40 & 0.78 & 0.80 & 0.56 & 0.75 & 0.66 & 0.76 & 0.42 & 0.74 & 0.07 & 0.93 \\
        Ours-cl-turn3 & 0.80 & 0.63 & 0.33 & 0.74 & 0.90 & 0.53 & 0.75 & 0.73 & 0.62 & 0.67 & 1.00 & 0.74 & 0.38 & 0.81 & 0.07 & 1.00 \\
        \hline
        Ours-cl-turn4 & 0.80 & 0.75 & 0.41 & 0.74 & 0.85 & 0.40 & 0.70 & 0.76 & 0.68 & 0.67 & 1.00 & 0.79 & 0.27 & 0.85 & 0.15 & 1.00 \\
        \hline
    \end{tabular}
    }
    \vspace{0.5em}
    \resizebox{\textwidth}{!}{
    \begin{tabular}{c|ccccc|ccccccccccc}
        \hline
        \multirow{2}{*}{\textbf{\raisebox{-1.0\height}{Method}}} & \multicolumn{5}{c|}{\textbf{Seen Categories}} & \multicolumn{11}{c}{\textbf{Unseen Categories}}\\
        \cline{2-17}
        & \includegraphics[width=0.050\linewidth]{./images/icon/folding_chair.png}
        &\includegraphics[width=0.050\linewidth]{./images/icon/toaster.png}
        &\includegraphics[width=0.050\linewidth]{./images/icon/lamp.png}
        &\includegraphics[width=0.050\linewidth]{./images/icon/dispenser.png}
        &\textbf{\raisebox{0.5\height}{AVG}}
        &\includegraphics[width=0.050\linewidth]{./images/icon/toilet.png}
        &\includegraphics[width=0.050\linewidth]{./images/icon/scissors.png}
        &\includegraphics[width=0.050\linewidth]{./images/icon/table.png}
        &\includegraphics[width=0.050\linewidth]{./images/icon/stapler.png}
        &\includegraphics[width=0.050\linewidth]{./images/icon/kettle.png}
        &\includegraphics[width=0.050\linewidth]{./images/icon/usb.png}
        &\includegraphics[width=0.050\linewidth]{./images/icon/switch.png}
        &\includegraphics[width=0.050\linewidth]{./images/icon/washing_machine.png}
        &\includegraphics[width=0.050\linewidth]{./images/icon/faucet.png}
        &\includegraphics[width=0.050\linewidth]{./images/icon/phone.png}
        &\textbf{\raisebox{0.5\height}{AVG}} \\
        \hline\hline 
        Ours-cl-turn1 & 0.00 & 0.56 & 0.29 & 0.60 & 0.61    & 0.00 & 0.26 & 0.35 & 0.00 & 0.71 & 0.26 & 0.65 & 0.53 & 0.12 & 0.62 & 0.32 \\
        Ours-cl-turn2 & 0.00 & 0.56 & 0.33 & 0.80 & 0.61 & 0.00 & 0.29 & 0.52 & 0.00& 0.50 & 0.45 & 0.56 & 0.43 & 0.17 & 1.00 & 0.40 \\
        Ours-cl-turn3 & 0.00 & 0.62 & 0.43 & 0.80 & 0.65 & 0.18 & 0.33 & 0.57 & 0.50 & 0.64 & 0.53 & 0.65 & 0.68 & 0.27 & 0.93 & 0.49 \\
        \hline
        Ours-cl-turn4 & 0.20 & 0.57 & 0.46 & 1.00 & 0.66 & 0.18 & 0.44 & 0.56 & 0.33 & 0.71 & 0.57 & 0.65 & 0.68 & 0.25 & 0.93 & 0.51 \\
        \hline
    \end{tabular}
    }
    \vspace{-0.1cm}
    \vspace{-0.1cm}
\end{table*}

The results presented in Table \ref{tab:testB} highlight the robustness and adaptability of our SC-VLA in a continual learning context. After four iterations of updates, our method does not overfit the Test set and demonstrates significant performance improvements across various object categories in the Test B set. Notably, our approach shows superior performance in categories such as \textit{Laptop (LT)}, \textit{Phone (PH)}, and \textit{Refrigerator (RF)}. This consistency underscores our model's ability to mitigate catastrophic forgetting effectively, ensuring sustained high performance through the integration of the Exponential Moving Average and fine-tuning process. 

\begin{table*}[htb]
\caption{\label{aptab:close}\textbf{The close-loop experiments.}  ``Fast" and ``Slow" represent our method's results for the fast system's direct pose prediction and the slow system's corrected prediction based on expert prompts, respectively. ``Mean" represents the average manipulation success rate across all tasks.}
\small
\vspace{-0.2cm}
\centering
\setlength\tabcolsep{1.1pt}
\begin{tabular}{c|c|c|c|c|c|c|c}
\toprule

  Method & Ref & Take USB out & Close fridge & Close box & Toilet seat down& Unplug charger & Mean \\\midrule
 ManipLLM & CVPR2024 & 0.45 & 0.90 & 0.55 & 0.95 & 0.55 & 0.68 \\
Ours(Fast) & - & 0.55 & 0.70 & 0.40 & 0.90 & 0.60 & 0.63 \\
Ours(Fast+Slow) & - & 0.85 & 0.95 & 0.75 & 1.0 & 1.0 &0.91\\\hline
\end{tabular}
\end{table*}

\section{Details of close-loop experiments}
\label{APsec:close}

In the "Main Results on Manipulation" section, we validate that SC-VLA can effectively perform both open-loop and closed-loop control in multiple tasks. In this section, we provide the detailed settings and experiment table for the closed-loop experiments. Specifically, we use the well-known closed-loop benchmark, RLBench~\citep{james2020rlbench}, as our closed-loop dataset. Five representative tasks with different manipulation trajectories were selected, including ``take USB out of computer," ``close fridge," ``close box," ``toilet seat down," and ``unplug charger." 
The evaluation metrics follow previous works \citep{james2020rlbench}, assessing task success rate based on predefined success conditions.
For the model input, we utilize the front-view image from a third-person perspective, as this view fully captures the manipulated object without occlusion. The key frame filtering method follows the previous work~\cite{shridhar2023perceiver}. We use 7-DoF to control the Franka Panda robot arm, including not only the position and rotation of the end-effector but also the opening and closing state of the gripper.
As shown in Table~\ref{aptab:close}, we compare our proposed method with the previous MLLM-based SOTA method~\citep{li2023manipllm} under the same experimental setting. Ours (Fast + Slow) achieves an average success rate of 91\%, improving accuracy by 23\% compared to ManipLLM. The results demonstrate that our slow system can also perform corrections for closed-loop control, showcasing the generalizability of our method.

\section{Additional related work}
\label{secap:rw}
In this section, we present additional related work that could not be fully discussed in the main text, highlighting the broader research landscape and its relation to our work. Specifically, we examine recent advances in multimodal large language models and techniques for robotic failure correction. These discussions underscore the strengths and limitations of existing approaches, motivating our development of a slow system that not only corrects errors but also continuously learns from corrective feedback.

\textbf{Multimodal Large Language Models.}  
Large language models (LLMs) have shown impressive reasoning capabilities across downstream tasks~\cite{touvron2023llama, floridi2020gpt}. In multi-modal reasoning, Multimodal Large Language Models (MLLMs)~\cite{alayrac2022flamingo, li2022blip, zhang2023llama} excel in bridging modalities.  
BLIP~\cite{li2022blip, li2023blip2} pre-trains encoder-decoder models with image-text pairs, enhancing vision-language understanding by adding synthetic captions and filtering noise. LLaVA~\cite{liu2023improved} and MiniGPT-4~\cite{zhu2023minigpt} introduce simple fully connected layers between image encoders and LLMs, emphasizing dataset pre-training and instruction tuning. Some MLLMs~\cite{chen2023shikra, wang2023visionllm, lin2023sphinx, wang2023cloud} focus on vision-centric tasks, confirming MLLMs' capacity for fine-grained perceptual results. Additionally, 3D MLLMs~\cite{guo2023point, hong20233d, wang2023chat, yang2023lidar} extend reasoning and conversational abilities to 3D modalities.
\begin{figure}[t]
\centering
   \includegraphics[width=0.95\columnwidth]{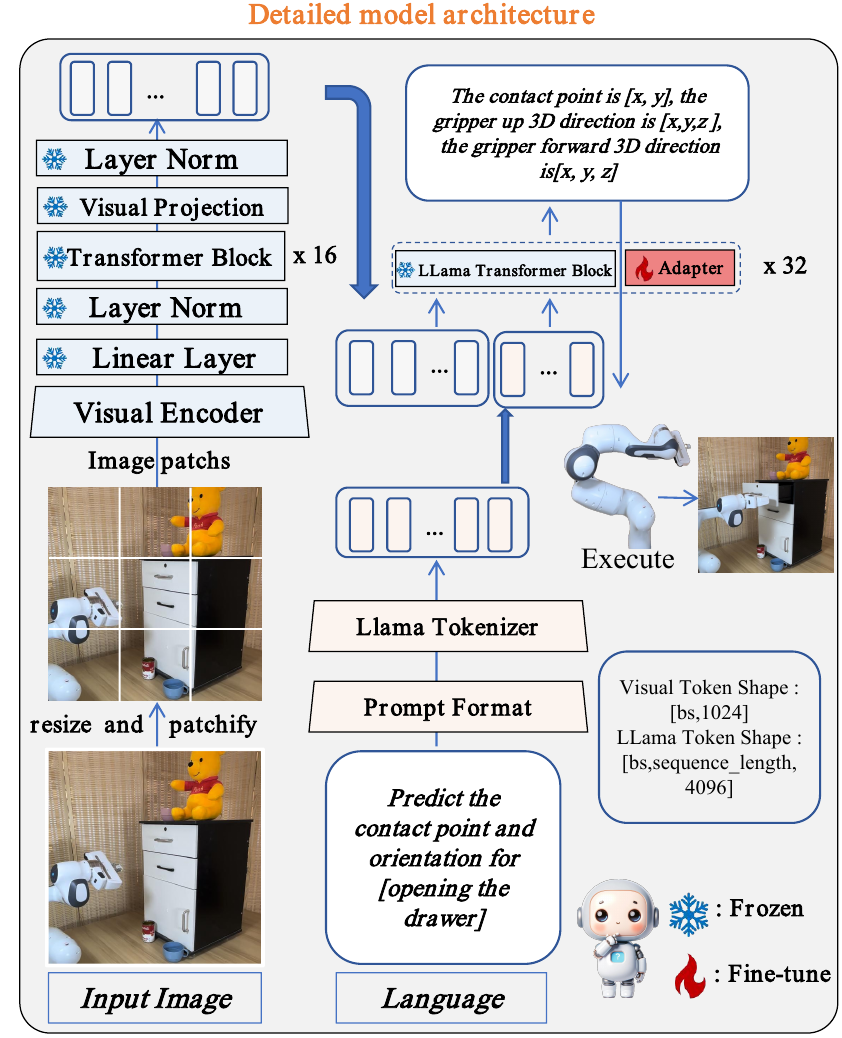}
\vspace{-0.2cm}
\caption{
\textbf{Detail architecture of SC-VLA.} }
\label{fig:detail}
\end{figure}

\textbf{Robotic Failure Correction.}  
While these approaches—namely REFLECT~\cite{liu2023reflect}, MULTIREACT~\cite{yu2023multireact}, DoReMi~\cite{guo2023doremi}, and CLAIRify~\cite{skreta2023errors} effectively address low-level execution failures, they do so by directly correcting immediate errors (such as 6 DoF end-effector poses) without incorporating the corrective feedback into a learning framework. REFLECT leverages LLMs to reason from summaries of past experiences, using failure explanations to refine planning; MULTIREACT employs a vision-language model~\cite{radford2021learning} as a reward mechanism to identify and autonomously recover from intermediate execution failures; DoReMi focuses on promptly detecting misalignments between planned and executed actions and then recovering from these discrepancies; and CLAIRify uses iterative prompting combined with program verification to ensure the validity of action plans.
These approaches, as mentioned above, are adept at handling immediate, low-level corrections but lack the ability to learn from the corrective interventions they perform. This limitation undermines their long-term adaptability and practical deployment in dynamic, real-world settings. In contrast, our work aims to develop a slow system that enables multi-modal large language models (MLLMs) not only to autonomously correct end-effector poses but also to continually incorporate and learn from the corrective feedback, thereby enhancing robustness and overall performance over time.

\section{Model architecture of SC-VLA}
\label{secap:ma}
As shown in Figure \ref{fig:detail} , to equip our model with foundational reasoning abilities for robotic manipulation, we load pretrained parameters from LLaMA-Adapter V2~\cite{gao2023llama}. 
This choice provides flexibility, as our approach can easily accommodate other MLLMs as the base model. As illustrated in Figure 2(in the main text), the model architecture is composed of several key components. When presented with an RGB image, we use the CLIP visual encoder~\cite{radford2021learning} to extract visual features. Simultaneously, a language prompt is processed by the pre-trained LLaMA tokenizer~\cite{touvron2023llama} to produce textual embeddings. To enable multimodal comprehension, our MLLM employs a projection layer to align the visual tokens from CLIP with the token embeddings of the LLaMA language model. This alignment allows the LLaMA model to understand both visual and textual information, thereby generating corresponding answers, such as the predicted end-effector pose for manipulation tasks. During training, we fine-tune only the injected adapters~\cite{hu2021lora} within the LLaMA model, keeping the majority of the pre-trained parameters frozen. This strategy preserves the robust capabilities of the pre-trained MLLM while enhancing the model with additional functionalities specific to manipulation and failure correction tasks. By injecting task-specific knowledge through adapters, we ensure that our model remains efficient without the need for large-scale retraining.

\section{Advantages of fast \& slow systems}
The dual-process structure in our model offers significant advantages in balancing speed, efficiency, and robustness for robotic manipulation tasks. By decomposing the action sequence into high-frequency predictions managed by the fast system and low-frequency corrections by the slow system, the model achieves an optimal allocation of computational resources. This separation enables rapid, direct pose predictions essential for responsive manipulation, while the slow system selectively intervenes to ensure accuracy and robustness, particularly under uncertain conditions.

Mathematically, the action sequence $X$ can be represented as:
\begin{equation}
    X = X_{\text{fast}} + X_{\text{slow}},
\end{equation}
where $X_{\text{fast}}$ handles high-frequency, immediate responses, and $X_{\text{slow}}$ captures the corrective actions. This decomposition allows the model to allocate more resources to the fast system for most interactions, activating the slow system only when necessary. This approach effectively reduces computational load and improves response times without compromising accuracy.

In robotic manipulation, entropy reflects SC-VLA's uncertainty regarding a specific action or prediction. This design's key advantage is its ability to mitigate cumulative errors by reducing the overall uncertainty of the model's predictions through decreased entropy. The slow system specifically targets and corrects high-entropy actions, effectively reducing the entropy $H(X)$ of the action sequence:
\begin{equation}
    H(X) = H(X_{\text{fast}}) + H(X_{\text{slow}}),
\end{equation}
where $H(X_{\text{slow}})$ represents the entropy reduction achieved through corrective interventions. This entropy minimization improves the reliability of predictions by reducing the risk of error propagation, especially in complex or novel manipulation tasks.

Furthermore, the fast system benefits from an adaptive Exponential Moving Average (EMA) update mechanism, incorporating corrections from the slow system to continuously improve performance. The EMA update is given by:
\begin{equation}
    \pi_t = \alpha \pi_{t-1} + (1 - \alpha) \pi_{\text{corr}},
\end{equation}
where $\alpha$ is dynamically adjusted based on task complexity or recent feedback. This adaptive weighting allows the model to prioritize recent corrections, enhancing the accuracy of future predictions across varying manipulation conditions. As a result, the model maintains high performance on both seen and unseen scenarios without extensive retraining.

The hierarchical design also supports an efficient approach to task execution, where the fast system generates initial, coarse predictions, and the slow system provides precision corrections as needed. This layered strategy improves the success rate of tasks and enables scalability to new scenarios, as evidenced by significant performance gains (see Table 1 in the main text). The dual-process model thus offers an effective balance of speed, precision, and adaptability, demonstrating its robustness in complex, real-world robotic applications.

\end{document}